\documentclass[10pt,twocolumn,letterpaper]{article}

\usepackage[pagenumbers]{cvpr} % To force page numbers, e.g. for an arXiv version

\newcommand{\vp}{\mathbf{p}}
\newcommand{\vt}{\mathbf{t}}

\newcommand{\vv}{\mathbf{v}}
\newcommand{\vz}{\mathbf{z}}
\newcommand{\vx}{\mathbf{x}}

\newcommand{\mE}{\mathbf{E}}
\newcommand{\mS}{\mathbf{S}}
\newcommand{\mI}{\mathbf{I}}

\newcommand{\ours}{SoC}

\usepackage[dvipsnames]{xcolor}
\usepackage{color, colortbl}

\definecolor{Gray}{gray}{0.9}

\newcommand{\our}{\cellcolor{Gray}}

\newcommand\blfootnote[1]{%
  \begingroup
  \renewcommand\thefootnote{}\footnote{#1}%
  \addtocounter{footnote}{-1}%
  \endgroup
}

\definecolor{cvprblue}{rgb}{0.21,0.49,0.74}
\usepackage[pagebackref,breaklinks,colorlinks,allcolors=cvprblue]{hyperref}

\usepackage{multirow}
\usepackage{arydshln}
\usepackage{amsthm}
\usepackage{bbm}
\usepackage{amsmath}
\usepackage{multicol}

\DeclareMathOperator*{\argmax}{arg\,max}

\newtheorem{proposition}{Proposition}

\newtheorem{corollary}{Corollary}
\theoremstyle{remark}

\definecolor{Better}{rgb}{0.18, 0.407, 0.266}
\definecolor{Worse}{rgb}{0.55, 0.22, 0.22}
\newcommand{\imp}[1]{$_{{\textbf{\textcolor{Better}{#1}}}}$}
\newcommand{\wor}[1]{$_{{\textbf{\textcolor{Worse}{#1}}}}$}

\newcommand{\mypar}[1]{\noindent\textbf{#1}}

\title{SoC: Semantic Orthogonal Calibration for Test-Time Prompt Tuning}

\author{Leo Fillioux\textsuperscript{1*} \and Omprakash Chakraborty\textsuperscript{2} \and Ismail Ben Ayed\textsuperscript{2} \and Paul-Henry Cournède\textsuperscript{1} \and Stergios Christodoulidis\textsuperscript{1} \and Maria Vakalopoulou\textsuperscript{1} \and Jose Dolz\textsuperscript{2}
\\
\and \textsuperscript{1} MICS, CentraleSupélec, Université Paris-Saclay, France
\\
\textsuperscript{2} LIVIA, ILLS, ÉTS Montréal, Canada
}

\begin{document}
\maketitle
\begin{abstract}
With the increasing adoption of vision-language models (VLMs) in critical decision-making systems such as healthcare or autonomous driving, the calibration of their uncertainty estimates becomes paramount. Yet, this dimension has been largely underexplored in the VLM test-time prompt-tuning (TPT) literature, which has predominantly focused on improving their discriminative performance. Recent state-of-the-art advocates for enforcing full orthogonality over pairs of text prompt embeddings to enhance separability, and therefore calibration. Nevertheless, as we theoretically show in this work, the inherent gradients from fully orthogonal constraints will strongly push semantically related classes away, ultimately making the model overconfident. Based on our findings, we propose Semantic Orthogonal Calibration (\textbf{\ours{}}), a Huber‑based regularizer that enforces smooth prototype separation while preserving semantic proximity, thereby improving calibration compared to prior orthogonality‑based approaches. Across a comprehensive empirical validation, we demonstrate that \ours{} consistently improves calibration performance, while also maintaining competitive discriminative capabilities.\blfootnote{\textsuperscript{*} Work done during an internship at ILLS.}
\end{abstract}    
\section{Introduction}
\label{sec:intro}

\begin{figure}[t]
    \centering
    \includegraphics[width=1\linewidth]{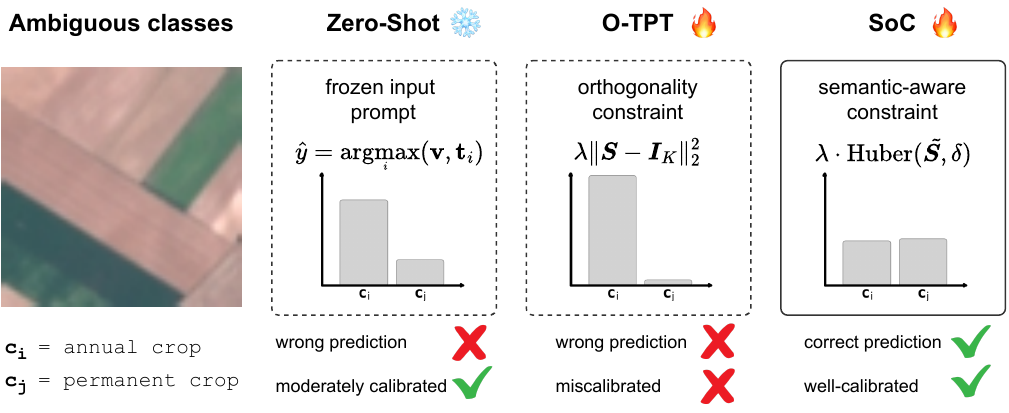} 
    \caption{\textbf{Motivation for SoC.} With O-TPT, ambiguity inherent to the class semantics is lost due to the aggressive orthogonality constraint, leading to artificially high confidence, even when predictions are incorrect. Let us take this image as an example, whose correct class is ``\texttt{annual crop land}'', and whose closest semantic class across all categories is ``\texttt{permanent crop land}''. The zero-shot CLIP prediction incorrectly classifies the image, but its prediction remains uncertain, as the softmax for those two closely related categories remains close. In contrast, due to its orthogonality constraints, O-TPT~\cite{sharifdeen2025tpt} pushes the text class prototypes apart, making the model become more confident, even if the prediction is wrong. Our proposed \textbf{\ours{}} addresses this issue with a smoother orthogonality enforcement.}
    \label{fig:teaser}
\end{figure}

Pre-trained vision-language models (VLMs) \cite{radford2021clip,xudemystifying}, built on millions of image-text pairs, have shown strong potential for capturing broad visual semantics and enabling generalization across a wide range of downstream vision tasks \cite{radford2021clip}. Yet, when deployed in real-world environments, these models inevitably confront the challenge of out-of-distribution data, for instance, in the form of tasks involving novel, unseen categories or domain shifts. This mismatch often undermines model generalization and scalability, particularly in scenarios where labeled data on the new task is scarce.  

A simple solution consists in leveraging the zero-shot transferability of VLMs through carefully hand-crafted textual prompts, such as ``\texttt{a photo of a [CLASS]}", which does not require labeled data of the target class. While effective, manual prompt design often relies on domain-specific heuristics and may fail to generalize across diverse tasks. To address this drawback, test-time prompt tuning (TPT) \cite{shu2022test} has emerged as a compelling alternative, enabling the optimization of textual prompts at inference without requiring labeled data and retraining. In particular, TPT learns prompt vectors via gradient descent, adaptively refining them using solely unlabeled test samples. By minimizing the entropy of the prediction distribution as a self-supervised signal, TPT allows VLMs, such as CLIP \cite{radford2021clip}, to better align with novel tasks and unseen class semantics at test time. Nevertheless, its reliance on entropy minimization as the only objective function induces overconfident predictions \cite{murugesan2024robust,yoon2024ctpt}, raising critical concerns about the reliability of VLMs. This is particularly important given the growing adoption of VLMs in real-world, safety-critical decision systems such as healthcare \cite{silva2025few,koleilat2025biomedcoop}, autonomous vehicles \cite{pan2024vlp,khandelwal2022simple}, or video-surveillance \cite{li2023clip}, where the calibration of their uncertainty estimates becomes crucial.  
 
To mitigate these limitations, calibration-oriented variants of TPT have been proposed, notably C-TPT \cite{yoon2024ctpt} and O-TPT \cite{sharifdeen2025tpt}. These approaches extend the original TPT paradigm by explicitly addressing the overconfidence induced by entropy minimization, encouraging dispersion between pairwise textual features. C-TPT \cite{yoon2024ctpt} introduces a regularization objective that spreads textual embeddings away from their centroid. Nevertheless, it treats dispersion as a proxy for calibration, which limits the control over prototype geometry and semantic coherence. O-TPT \cite{sharifdeen2025tpt}, in contrast, operates directly on the geometry of the class prototype manifold, enforcing pairwise orthogonality between the class text embeddings. Although this improves separation, the strong repulsion for semantically similar (i.e., collinear) prototypes may actually be detrimental when the similarity reflects meaningful semantic overlap. For example, classes like \textit{dog} and \textit{puppy} are expected to be close in both image and text embedding spaces. Forcing such prototypes toward orthogonality disrupts the learned manifold and risks over-separating classes that should remain geometrically close. Furthermore, as we will demonstrate both analytically and empirically, the strong repulsion derived from full orthogonality results in systematic increases in confidence, which deteriorates model calibration, particularly for semantically similar categories. 
Based on these observations, the three key contributions of this study are:
\begin{itemize}
    \item We introduce Semantic Orthogonal Calibration (SoC), a  Huber-based regularizer for TPT, which yields smoother gradients than the full orthogonality constraints used in prior work (Fig.~\ref{fig:teaser}). While both induce systematic distortions in the embedding space, our formulation caps the repulsion for highly similar pairs, thereby better preserving semantic structure among related classes and mitigating the over‑separation effects of strict orthogonality.
    \item We derive a \emph{lower bound on the confidence} that links worst-case class similarity $\mu$ to prediction uncertainty, revealing how geometric repulsion influences calibration. This analysis shows that full orthogonality aggressively reduces $\mu$, even for semantically correlated categories, leading to high confidence increase and degraded calibration. In contrast, the proposed regularizer applies gentler repulsion to high-similarity pairs, avoiding excessive confidence increases and better preserving semantic proximity, which ultimately yields better calibration. 
    \item Comprehensive experiments across diverse benchmarks on fine-grained classification and natural domain shifts demonstrate the superiority of our approach. \ours{} consistently improves calibration metrics over state-of-the-art baselines, while maintaining highly competitive discriminative performance, showing strong generalization across different backbones and diverse initial text prompts. 
\end{itemize}
\section{Related Work}
\label{sec:related}

\mypar{Prompt tuning vision-language models.}
Vision-language models were introduced with CLIP~\cite{radford2021clip} as a way of jointly training a vision and textual encoder to encode image-text pairs in a joint $\ell_2$-normalized space. These models are trained in a contrastive manner to maximize cosine similarity between positive pairs while minimizing it for negative pairs. The following works have built upon this pipeline, and explored various aspects such as scaling the dataset (\cite{jia2021scaling}), exploring architectural choices and feature space fusion (\cite{ALBEF, mu2022slip}), and the integration with large language models (\cite{alayrac2022flamingo, li2022blip, liu2023llava}). One important feature of VLMs is their strong zero-shot performance, which enables them to classify input images given a text prompt and the class names. The image is classified using the highest cosine similarity between its embedding and the class embeddings. Prompt tuning allows to adapt the input prompt of a frozen VLM to a target task, using a limited number of samples (i.e., few-shot setting). CoOp~\cite{zhou2022coop} was introduced as a way to replace hand-crafted prompts with learnable prompts, which was later extended with CoCoOp~\cite{zhou2022cocoop} to include input-adaptive prompts. KgCoOp~\cite{kgcoop23} builds upon CoOp, improving its generalizability to new classes by reducing the discrepancy between the hand-crafted and the learned prompt. Test-time prompt tuning (TPT)~\cite{shu2022tpt} shows that learning the input prompt at test-time (i.e., zero-shot setting) can increase the performance of the zero-shot classification in VLMs. Given an input test image and applying a batch of augmented views, the input prompt is updated in a single step of gradient descent by using the cross-entropy of the average prediction.

\noindent \textbf{Calibrating modern neural networks.} Model calibration has gained increasing popularity, where mainly post-processing \cite{guo2017calibration,yang2023beyond,zhang2020mix} and training-time \cite{pereyra2017regularizing,muller2019does,mukhoti2020calibratingSup,liu2022devil,liu2023class} approaches have emerged. In particular, temperature scaling \cite{guo2017calibration} and its enhanced adaptive variants \cite{yang2023beyond,zhang2020mix} artificially increase the entropy of network predictions post hoc, whereas training-based methods typically aim to maximize the entropy of softmax outputs either explicitly \cite{pereyra2017regularizing,cheng2022calibrating} or implicitly \cite{muller2019does,mukhoti2020calibratingSup} during training. Some other training-based approaches further incorporate penalty-based objectives into the primary loss function to regulate the separation between logits \cite{liu2022devil,liu2023class}. Furthermore, data augmentation methods, such as Mixup \cite{zhang2018mixup}, CutMix \cite{yun2019cutmix}, and AugMix \cite{hendrycks2020augmix}, train deep models on mixed samples to mitigate overconfident predictions.

\noindent \textbf{Calibrating vision-language models.} With the rapid rise of VLMs and their growing adoption in safety-critical applications, recent works have begun to explicitly address the uncertainty in their predictions. In particular, most of these approaches have focused on either full fine-tuning \cite{wang2024open,tu2024empirical,oh2024towards,lv2025contrast} or few-shot learning \cite{morales2025bayesadapter} settings, which both require labeled samples, differing from the scenario studied in this work. SaLS \cite{murugesan2024robust} exposed that while most of the adaptation approaches enhance accuracy, they do it at the cost of degrading zero-shot calibration, and proposed an unsupervised logit normalization strategy to calibrate VLMs across different labeled regimes, including test-time prompt tuning. Closely related to our work, C-TPT~\cite{yoon2024ctpt} introduces a dispersion-based loss that encourages text embeddings to spread away from the class centroids, enhancing inter-class separability. While this improves diversity, it may underperform in complex scenarios due to limited utilization of the embedding space. O-TPT~\cite{sharifdeen2025tpt} builds on these weaknesses and enforces full orthogonality between the prompt class prototypes. In contrast, our approach introduces a smoother regularization mechanism that respects semantic proximity, avoiding the rigid separation imposed by full orthogonality, yet making more efficient use of the embedding space than dispersion-based methods like C-TPT.
\section{Preliminaries}
\label{sec:method}

We first formally define the setting for VLM zero-shot classification before introducing the baseline methods.

\subsection{Problem setting}
A pretrained VLM consists of a vision encoder $f_{\boldsymbol{\theta}}(\cdot)$ and a textual encoder $f_{\boldsymbol{\phi}}(\cdot)$, which generate the image embedding $\vv = f_{\boldsymbol{\theta}}(\vx)\in\mathbb R^d$ and the class prototypes $\vt_k \in \mathbb R^d$, respectively. The class prototypes are generated by feeding a text template to the text encoder, $\vt_k=f_{\boldsymbol{\phi}}(\texttt{"a photo of a [CLASS]"})$. Note that we resort to CLIP as a prominent VLM in the literature, which embeds the image and text embeddings in an $\ell_2$-normalized space, i.e. $\Vert \vv\Vert=\Vert\{\vt_k\}_{k=1}^K\Vert = 1$, with $K$ being the number of classes. Let us define the logits and softmax probabilities:
\begin{equation}
\vz_k = \alpha \, \vv^\top \vt_k, 
\qquad 
\vp_k(\vv) = \frac{\exp(\vz_k)}{\sum_{j=1}^K \exp(\vz_j)},\nonumber
\end{equation}
\noindent where $\vz=(z_k)_{1 \leq k \leq K}$, $\vp=(p_k)_{1 \leq k \leq K}$, and $\alpha = \tfrac{1}{T}$, with $T$ being the temperature scaling value controlling the shape of the softmax distributions, learned during pretraining \cite{radford2021clip}.

We define $\mE \in \mathbb R^{K\times d}$ as the matrix that contains all class prototypes. In addition, given that prototypes are unit norm, $\mS= \mE \mE^\top$ defines the cosine similarity, where $s_{ij}~=~\vt_i^\top\vt_j$
is the cosine similarity between the class prototypes of classes $i$ and $j$.

\subsection{Relevant approaches}

\mypar{TPT~\cite{shu2022tpt}.} Test-Time Prompt Tuning (TPT) optimizes the input text prompts by resorting to a cross entropy loss:
\begin{equation}
    \mathcal L_\text{TPT} = -\sum_{k=1}^K \tilde{p}_k(\vv) \log \tilde{p}_k(\vv),\nonumber
\end{equation}
\noindent where, for a given test image $\vx$, and its corresponding embedding $\vv$, $\tilde{p}_k(\vv)$ is the average softmax probability across $N$ augmentations, for the $k$-th class, thresholding to only keep the most confident predictions:
\begin{equation}
\tilde{p}_k(\vv)=\frac{1}{\rho N}\sum_{n=1}^N \mathbbm 1(\mathcal H(p_k) \geq \tau) p_k(\mathcal A_n(\vv)),\nonumber
\end{equation}
with $\mathcal A_n(\vv)$ corresponding to the visual embedding of the image augmented with the $n$-th augmentation, and $\tau$ is the $\rho$-percentile of the entropy $\mathcal H$ of $N$ augmented views ranked from low to high.

\mypar{C-TPT~\cite{yoon2024ctpt}.} Based on TPT, C-TPT introduces a regularization term to maximize the distance between the text prototypes and their centroid:
\begin{equation}
    \mathcal L_\text{C-TPT} = \mathcal L_\text{TPT} - \lambda \cdot \frac{1}{K}\sum_{k=1}^K \Vert \bar{\vt} - \vt_k \Vert_2,\nonumber
\end{equation}
where $\bar{\vt}=\frac{1}{K}\sum_k\vt_k$ is the centroid of textual embeddings.

\mypar{O-TPT~\cite{sharifdeen2025tpt}.} In a similar approach, O-TPT regularizes the TPT loss by adding a term that forces full orthogonality across pairs of text class prototypes.
\begin{equation}
    \mathcal L_\text{O-TPT} = \mathcal L_\text{TPT} + \lambda \Vert \mS-\mI_K \Vert
    _2^2,\nonumber
\end{equation}
\noindent with $\mI_K$ denoting the identity matrix of dimension $K$, which removes the diagonal terms of the $\mS$ matrix.

\subsection{Limitations of O-TPT}
\begin{figure}
    \centering
    \includegraphics[width=\linewidth]{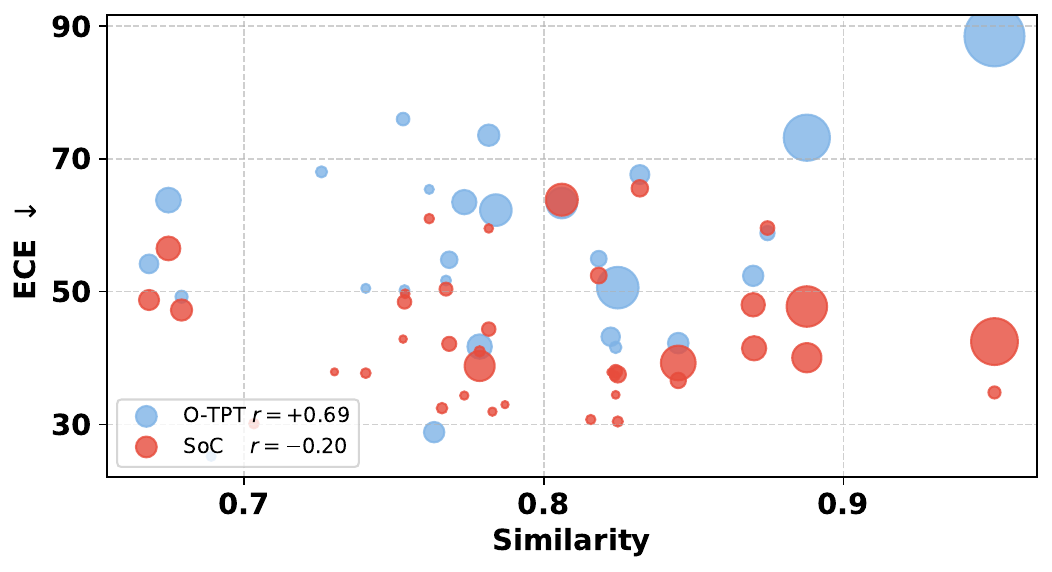}
    \caption{\textbf{ECE per class pair as a function of the zero-shot cosine similarity.} We compute the ECE for the wrong predictions across each class pair (i.e., the model predicted class $i$ when the label was class $j$) and analyze the relation with the zero-shot similarity between both classes on EuroSAT. For classes with high initial semantic similarity, O-TPT is overconfident, caused by the underlying drawbacks of enforcing orthogonality across all pairs. Circle size indicates the number of samples in each $(i, j)$ pair.
    }
    \label{fig:ece_sim}
\end{figure}

Despite the performance gains compared to C-TPT \cite{yoon2024ctpt}, O-TPT exhibits several limitations. In particular, full orthogonality imposes a quadratic penalty on pairwise similarity. This means that highly similar pairs are forced apart more aggressively than less similar ones. While this behaviour is initially sought, as it promotes separation, it can be suboptimal for semantically correlated classes, i.e., those that are expected to have geometrical proximity due to conceptual overlap. For instance, consider the classes \textit{annual crop land} and \textit{permanent crop land}. These categories are naturally close in both image and text embedding manifolds, and their similarity reflects meaningful semantic structure. Under full orthogonality, their high similarity triggers strong repulsion, pushing them apart even though their proximity is desirable. Fig.~\ref{fig:ece_sim} perfectly illustrates this, by showing how O-TPT makes overconfident predictions in classes that are semantically very close (i.e., with high zero-shot similarity).

\section{Our proposed SoC regularizer}
Building on the above observations, we propose to enforce class prototype separation in a smoother manner, with a regularizer that respects semantic proximity. Specifically, we resort to the Huber loss \cite{huber1964robust}, which is quadratic near zero but transitions to linear for larger residuals, preventing steep gradient growth for semantically similar pairs and making it a natural choice for our regularizer.

Given a margin $\delta\in\mathbb [0, 1]$, we define our prompt tuning regularizer as: 
\begin{equation}
    \mathcal L_\text{Huber}(s, \delta) =
    \begin{cases}
    \frac{s^2}{2} & \text{if } s \leq \delta \\
    \delta(s-\frac{\delta}{2}) & \text{otherwise,}
    \end{cases}\nonumber
\end{equation}
which is integrated into the whole learning objective:
\begin{equation}
\mathcal L_\text{\ours} = \mathcal L_\text{TPT} + \lambda\cdot \frac{2}{K(K-1)} \sum_{i<j}\mathcal L_\text{Huber}(s_{ij}, \delta),\nonumber
\end{equation}
with $\frac{K(K-1)}{2}$ denoting the number of elements in the lower triangle of the $\mS$ matrix.

\subsection{Confidence bound under cosine coherence}

We next formalize how the degree of prototype similarity directly controls the confidence floor of the softmax distribution. Let $i^\star = \arg\max_i z_i$, and define the softmax confidence as:
\begin{equation}
p_{\max}(\vv) = \max_i p_i(\vv) 
= \frac{1}{1 + \sum_{j \neq i^\star} \exp\!\big(-\Delta z_{i^\star j}(\vv)\big)},\nonumber
\end{equation}
where the logit gap can be formally defined as:
\begin{equation}
\Delta z_{i^\star j}(\vv) = z_{i^\star} - z_j = \alpha \, \vv^\top (\vt_{i^\star} - \vt_j).\nonumber
\end{equation}
Then, for any set of $K$ classes ($K \geq 2$), let us define the \textit{cosine coherence} of the set as:
\begin{equation}
\mu \triangleq \max_{i \neq j} \, \vt_i^\top \vt_j \in [0,1]\footnote{We assume CLIP-style class embeddings, which are unit-normalized and yield non-negative pairwise cosine similarities.}.\nonumber
\end{equation}

\begin{proposition} [Confidence floor via cosine coherence]
\label{prop:bound}
For any unit vector $\vv$, the confidence of the prediction satisfies
\begin{equation}
\label{eq:bound}
p_{\max}(\vv) \;\ge\; \frac{1}{1+(K-1)\exp(-\alpha(1-\mu))}.
\end{equation}
\end{proposition}
\noindent
The proof is deferred to the Appendix \Cref{sec:proof_prop}.

\subsection{First-order analysis}
\label{subsec:first_order}
To understand how orthogonality regularization influences model confidence and calibration, we analyze the immediate effect of a single gradient step on the geometry of the prototype space (note that O-TPT~\cite{sharifdeen2025tpt}, similarly to other TPT-based methods, performs only one gradient step over text embeddings). Specifically, we investigate how the worst-case similarity $\mu = \max_{i\ne j} s_{ij}$ evolves under full orthogonality (O-TPT) and the proposed Huber-style regularizer. This quantity directly controls the softmax confidence lower bound (Proposition~\ref{prop:bound}), and its reduction is often interpreted as a proxy for improved separation. However, the mechanism by which $\mu$ is reduced has important implications for semantic preservation and calibration.

Let us consider only a single gradient step of size $\eta>0$, use the first-order approximation, and assume the worst-case similarity $\mu = \max_{i \ne j} s_{ij}$ is attained by the dominant pair in the set. We now formalize the update dynamics. A single gradient step of size $\eta$ yields the updated prototypes $\vt_i' = \vt_i - \eta \nabla_{\vt_i}$, and updated similarity $s_{ij}' = (\vt_i')^\top \vt_j'$. Thus, the first-order pairwise similarity shift, which captures how the cosine similarity between class prototypes $\vt_i$ and $\vt_j$ evolves, can be defined as:
\begin{equation}
\label{eq:approx}
\Delta s_{ij} \equiv s_{ij}' - s_{ij} \approx -\eta (\vt_j^\top \nabla_{\vt_i} + \vt_i^\top \nabla_{\vt_j}),
\end{equation}
where the last approximation follows from a first-order Taylor expansion, with the $O(\eta^2)$ term ignored, as it is negligible under small-step first-order dynamics.

Full orthogonality leads to the following gradient, $\nabla_{\vt_i}^{\text{O-TPT}} = 2 \sum_{k \ne i} s_{ik} \vt_k$, whereas the gradient from the proposed Huber-style regularized is:
\begin{equation}
\nabla_{\vt_i}^{\text{Huber}} = \sum_{k \ne i} g_\delta(s_{ik}) \vt_k,
\;\text{where}\quad
g_\delta(s) =
\begin{cases}
s, & s \le \delta, \\
\delta, & s > \delta.
\end{cases}\nonumber
\end{equation}
For analytical clarity, we approximate the gradient by retaining only the dominant pair $(i, j)$, i.e., the neighbors attaining the highest similarity. Specifically:
\begin{equation}
\nabla_{\vt_i}^{\text{O-TPT}} \approx 2\, s_{ij}\, \vt_j,\nonumber
\qquad
\nabla_{\vt_j}^{\text{O-TPT}} \approx 2\, s_{ij}\, \vt_i,\nonumber
\end{equation}
and
\begin{equation}
\nabla_{\vt_i}^{\text{Huber}} \approx g_{\delta}(s_{ij})\,\vt_j,\nonumber
\qquad
\nabla_{\vt_j}^{\text{Huber}} \approx 2\, g_{\delta}(s_{ij})\,\vt_i.\nonumber
\end{equation}

Note that this simplification still highlights the repulsion effect without altering the qualitative behavior of the update (the exact derivation considering all pairs is derived in Appendix~\ref{sec:grad_allpairs}). Using the first-order update rule (Eq. \ref{eq:approx}):
\begin{align}
\label{eq:approx_otpt}
\Delta s_{ij}^{\text{O-TPT}} &\approx -\eta (\vt_j^\top\,(2s_{ij}\vt_j)  + \vt_i^\top \,(2s_{ij}\vt_j)),\nonumber\\ 
&\approx -2 \eta s_{ij}(\|\vt_j\|^2+\|\vt_i\|^2)\nonumber\\
& \approx -4 \eta s_{ij}.
\end{align}
Similarly, for the proposed Huber-based regularizer:
\begin{align}
\Delta s_{ij}^{\text{Huber}} &\approx -\eta (\vt_j^\top\,(g_{\delta}(s_{ij})\vt_j)  + \vt_i^\top \,(g_{\delta}(s_{ij})\vt_j)),\nonumber\\ 
&\approx -\eta \,g_{\delta}(s_{ij})(\|\vt_j\|^2+\|\vt_i\|^2)\nonumber\\
& \approx -2 \eta \, g_{\delta}(s_{ij})=
\begin{cases}
-2\eta\, s_{ij}, & s_{ij} \le \delta, \\
-2\eta\, \delta, & s_{ij} > \delta.
\end{cases} \label{eq:delta-huber}
\end{align}

Then, after one step, considering $\mu = s_{ij}$, and the gradients in Eq. (\ref{eq:approx_otpt}) and (\ref{eq:delta-huber}):
\begin{align}
\mu'_{\text{O-TPT}} &\approx
(1 - 4\eta)\, \mu,\nonumber \\
\mu'_{\text{Huber}} &\approx
\begin{cases}
(1 - 2\eta)\, \mu, & \mu \le \delta, \\
\mu - 2\eta\, \delta, & \mu > \delta.
\end{cases}\nonumber
\end{align}
The behavior of the proposed Huber regularizer depends on whether $\mu$ lies below or above the threshold 
$\delta$:
\begin{itemize}
    \item \noindent When \textit{\boxed{\mu \le \delta}}:
the Huber gradient yields a similarity update of $\mu'_{\text{Huber}} = (1 - 2\eta)\mu$. In contrast, O-TPT applies a stronger repulsion, resulting in $\mu'_{\text{O-TPT}} = (1 - 4\eta)\mu$. Thus, even in the low-similarity regime, O-TPT contracts $\mu$ more aggressively than Huber. 
\item \noindent When \textit{\boxed{\mu > \delta}}: the Huber update becomes capped ($\mu'_{\text{Huber}} = \mu - 2\eta \delta$), while O-TPT continues to scale with $\mu$ ($\mu'_{\text{O-TPT}} = (1 - 4\eta) \mu$). 
\end{itemize}
Comparing both, we find that:
\begin{equation}
\mu'_{\text{O-TPT}} < \mu'_{\text{Huber}} \quad \text{iff} \quad 4\mu > 2\delta,\nonumber
\end{equation}
\noindent which holds whenever $\mu > \delta$. This confirms that O-TPT consistently yields a sharper one-step reduction than the proposed Huber regularizer in worst-case similarity, regardless of the threshold $\delta$.

\begin{table*}[t]
    \caption{\textbf{Benchmark with the ViT-L/14 backbone.} Accuracy and ECE across 11 datasets compared with four baseline methods. Green and red indicate positive and negative changes with respect to state-of-the-art O-TPT \cite{sharifdeen2025tpt}, respectively.}
    \label{tab:vitl}
    \centering
    \resizebox{\linewidth}{!}{
    \begin{tabular}{l l c c c c c c c c c c c c}
        \toprule
         & & ImgNet
         & DTD
         & Flowers
         & Food101
         & SUN397
         & Aircraft
         & Pets
         & Caltech
         & UCF101
         & EuroSAT
         & Cars
         & Average \\
        \midrule
        \multirow{5}{*}{\rotatebox{90}{\textbf{Accuracy}}} & Zero-Shot &
            73.5 &
            52.4 &
            76.2 &
            88.6 &
            67.7 &
            29.9 &
            93.1 &
            95.1 &
            73.8 &
            55.0 &
            76.8 &
            71.1 \\
        & TPT\textcolor{gray}{$_\text{ NeurIPS'22}$} &
            75.6 &
            55.3 &
            76.3 &
            89.0 &
            70.2 &
            31.8 &
            93.6 &
            95.5 &
            74.9 &
            51.9 &
            77.8 &
            72.0 \\
        & C-TPT\textcolor{gray}{$_\text{ ICLR'24}$} &
            75.0 &
            55.1 &
            76.5 &
            88.9 &%
            70.1 &
            30.9 &
            94.1 &
            95.5 &
            75.2 &
            54.0 &
            77.5 &
            72.1 \\
        & O-TPT\textcolor{gray}{$_\text{ CVPR'25}$} &
            73.2 &
            54.6 &
            76.4 &
            88.6 &
            68.9 &
            30.0 &
            93.8 &
            95.3 &
            74.5 &
            53.6 &
            76.7 &
            71.4 \\
        & \our \textbf{\ours{} (\textit{Ours})} &
            \our74.5\imp{+1.3} &
            \our54.4\wor{-0.2} &
            \our77.0\imp{+0.6} &
            \our88.9\imp{+0.3} &
            \our69.5\imp{+0.6} &
            \our30.9\imp{+0.9} &
            \our93.9\imp{+0.1} &
            \our95.6\imp{+0.3} &
            \our74.9\imp{+0.4} &
            \our58.3\imp{+4.7} &
            \our77.0\imp{+0.3} &
            \our72.3\imp{+0.9} \\
        \midrule
        \multirow{5}{*}{\rotatebox{90}{\textbf{ECE}}} & Zero-Shot &
            2.9 &
            10.3 &
            4.3 &
            1.8 &
            3.3 &
            10.3 &
            2.8 &
            3.7 &
            5.6 &
            6.9 &
            3.9 &
            5.1 \\
        & TPT\textcolor{gray}{$_\text{ NeurIPS'22}$} &
            14.8 &
            25.0 &
            15.0 &
            6.2 &
            17.2 &
            28.8 &
            3.7 &
            2.4 &
            17.7 &
            25.8 &
            7.1 &
            14.9 \\
        & C-TPT\textcolor{gray}{$_\text{ ICLR'24}$} &
            10.5 &
            18.3 &
            11.1 &
            3.3 &
            13.5 &
            21.4 &
            1.0 &
            0.9 &
            11.1 &
            17.3 &
            1.4 &
            10.0 \\
        & O-TPT\textcolor{gray}{$_\text{ CVPR'25}$} &
            5.5 &
            13.8 &
            7.0 &
            2.8 &
            7.6 &
            16.8 &
            1.4 &
            2.0 &
            8.5 &
            17.7 &
            2.2 &
            7.7 \\
        & \our \textbf{\ours{} (\textit{Ours})} &
            \our7.2\wor{+1.7} &
            \our10.9\imp{-2.9} &
            \our5.3\imp{-1.7} &
            \our2.1\imp{-0.7} &
            \our7.2\imp{-0.4} &
            \our12.7\imp{-4.1} &
            \our0.5\imp{-0.9} &
            \our1.3\imp{-0.7} &
            \our6.5\imp{-2.0} &
            \our3.2\imp{-14.5} &
            \our2.2\imp{-0.0} &
            \our5.4\imp{-2.3} \\
        \bottomrule
    \end{tabular}}
\end{table*}
\begin{corollary}[Confidence increases under full orthogonality]
Let $\mu = \max_{i \ne j} s_{ij}$ denote the worst-case similarity between class prototypes, and let $p_{\max}(\vv)$ be the softmax confidence lower bound as defined in Proposition~\ref{prop:bound}. Then, under a single gradient step of size $\eta$, full orthogonality (O-TPT~\cite{sharifdeen2025tpt}) yields a strictly lower worst-case similarity than Huber-style regularization:
\begin{equation}
\mu'_{\text{O-TPT}} < \mu'_{\text{Huber}},\nonumber
\end{equation}
and therefore a strictly higher confidence bound:
\begin{equation}
p_{\max}^{\text{O-TPT}}(\vv) > p_{\max}^{\text{Huber}}(\vv).\nonumber
\end{equation}
\vspace{-4mm}
\label{cor:confidence_increase}
\end{corollary}
This result highlights the geometric mechanism by which O-TPT inflates confidence more aggressively than the proposed Huber-based regularizer, \ours{}.
\section{Experiments}
\label{sec:experiments}

\subsection{Experimental setup}
\mypar{Datasets.} Building upon previous works~\cite{shu2022tpt,yoon2024ctpt,sharifdeen2025tpt}, we first evaluate our approach on a comprehensive benchmark that encompasses \textbf{11 diverse image classification datasets}, including: two generic-objects datasets (ImageNet \cite{5206848}, Caltech101 \cite{li2006one}), five fine-grained datasets (OxfordPets \cite{parkhi12a}, StanfordCars \cite{KrauseStarkDengFei-Fei_3DRR2013}, Flowers102 \cite{Nilsback08}, Food101 \cite{bossard14}, FGVC-Aircraft \cite{maji13fine-grained}), a scene recognition dataset (SUN397 \cite{xiao2010sun}), an action recognition dataset (UCF101 \cite{Soomro2012UCF101AD}), a texture dataset (DTD \cite{cimpoi14describing}), and a satellite image dataset (EuroSAT \cite{helber2017eurosat}). Furthermore, we also evaluate \ours{} on the four variants of the popular ImageNet, namely ImageNet-A \cite{imagenet_a}, ImageNet-v2 \cite{pmlr-v97-recht19a}, ImageNet-R \cite{hendrycks2021many}, and ImageNet-Sketch \cite{wang2019learning}. Further datasets details are presented in Appendix~\ref{sec:data}.

\noindent \textbf{Baselines.} We use TPT \cite{shu2022tpt} as the main baseline, whose learning objective is solely focused on improving classification accuracy. Furthermore, we include C-TPT \cite{yoon2024ctpt} and O-TPT \cite{sharifdeen2025tpt} as relevant concurrent methods that have been proposed to enhance TPT. We conduct the experiments under identical settings across methods to ensure a fair comparison.

\mypar{Implementation details.} We use ViTs of different sizes as backbone networks (ViT-L/14 and ViT-B/16, with ViT-L/14 used across experiments, unless otherwise stated), as they have been shown to be better calibrated than CNNs~\cite{NEURIPS2021_8420d359}, and thus represent a more realistic scenario. Following \cite{yoon2024ctpt,sharifdeen2025tpt}, prompts are initialized as ``\texttt{a photo of a [CLASS]}" and optimized using AdamW \cite{loshchilov2019decoupled} over a single gradient step, and a learning rate of 0.005. The batch size is set to 64 across all experiments (corresponding to 64 augmentations of each image), similar to \cite{sharifdeen2025tpt}. All remaining settings strictly adhere to the configuration in \cite{yoon2024ctpt,sharifdeen2025tpt}. Further details are provided in Appendix Section~\ref{sec:implementation}.

\noindent \textbf{Evaluation metrics.} To evaluate the classification performance, we rely on accuracy. Then, in order to compare the calibration with C-TPT and O-TPT, we report the expected calibration error (ECE). Given a test set of size $M$ with visual embeddings $V=\{\vv_1,\:\ldots,\:\vv_M\}$ with labels $Y=\{y_1,\:\ldots,\:y_M\}$, and class prototypes $T=\{\vt_1,\:\ldots,\:\vt_K\}$. Let us define by $\gamma(\vv,y) = \mathbbm 1\big(\argmax_j(\vv^\top\vt_j)=y  \big)$ the binary operator for determining correct sample-wise classification, and by $\epsilon(\vv, b)=\mathbbm 1\big (\text{softmax}(\max_j(\vv^\top \vt_j)) \in b\big )$ the binary operator indicating if the maximum softmax for $\vv$ falls in bin $b$.
\begin{equation}
    \text{ECE}(V,Y,T)=\frac{1}{\vert B \vert}\sum_{b_i\in B} n_i\vert \gamma(\vv_m, y_m)\cdot\epsilon(\vv_m, b_i) -\bar{b_i}\vert\nonumber
\end{equation}
\noindent
with $B$ the bins into which to split the confidence scores, $n_i=\sum_{\vv_m}\epsilon(\vv_m, b_i)$ the number of elements in the $i$-th bin, and $\bar{b_i}$ the median bin value.

\subsection{Results}
\noindent \textbf{Performance on \textit{fine-grained} classification tasks.} Tab.~\ref{tab:vitl} reports the results across 11 fine-grained classification datasets for the different approaches. We can observe that, while \ours{} improves the overall classification performance compared to previous TPT-based methods, it also yields the best calibrated prediction, whose mean ECE differences range from 2.3 (compared to O-TPT) to 9.5 with respect to the original TPT method. Particularly, compared to the previous state-of-the-art O-TPT, \ours{} achieves the best ECE scores in all but one dataset, showing its robustness across multiple datasets. Furthermore, it is noteworthy to mention that, from a calibration standpoint, our approach performs on par with zero-shot, which has been shown in recent literature to provide the best calibration~\cite{murugesan2024robust}.

\begin{figure}[h!]
    \centering
    \begin{subfigure}[b]{0.32\linewidth}
        \centering
        \includegraphics[width=\linewidth]{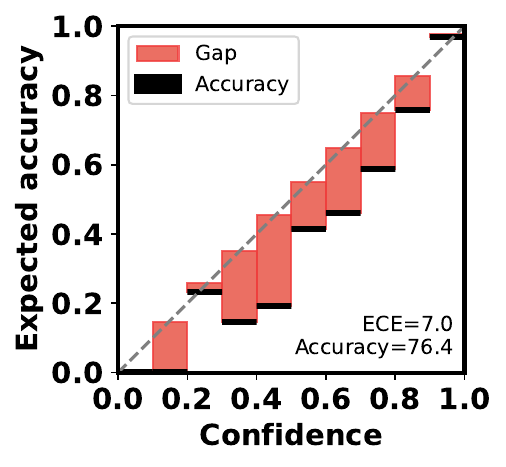}
        \caption{Flowers O-TPT}
    \end{subfigure}
    \begin{subfigure}[b]{0.32\linewidth}
        \centering
        \includegraphics[width=\linewidth]{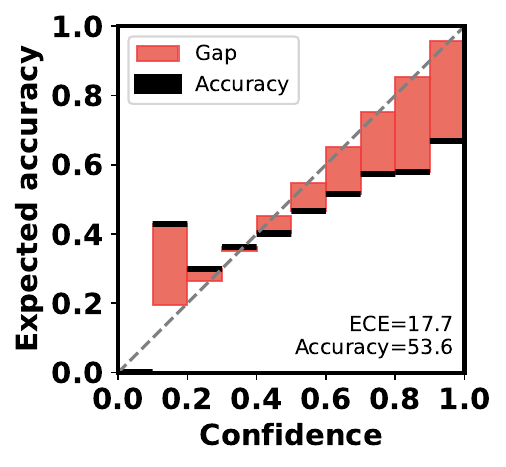}
        \caption{EuroSAT O-TPT}
    \end{subfigure}
    \begin{subfigure}[b]{0.32\linewidth}
        \centering
        \includegraphics[width=\linewidth]{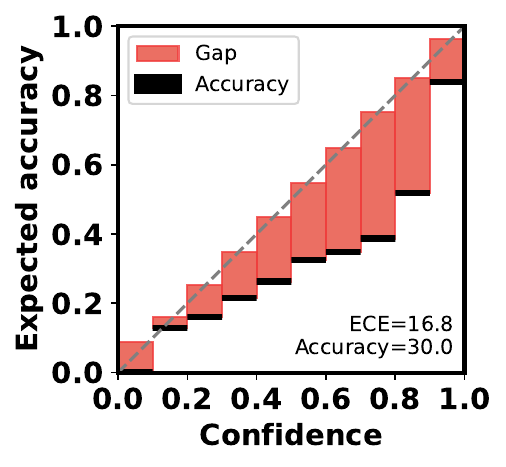}
        \caption{Aircraft O-TPT}
    \end{subfigure}

    \begin{subfigure}[b]{0.32\linewidth}
        \centering
        \includegraphics[width=\linewidth]{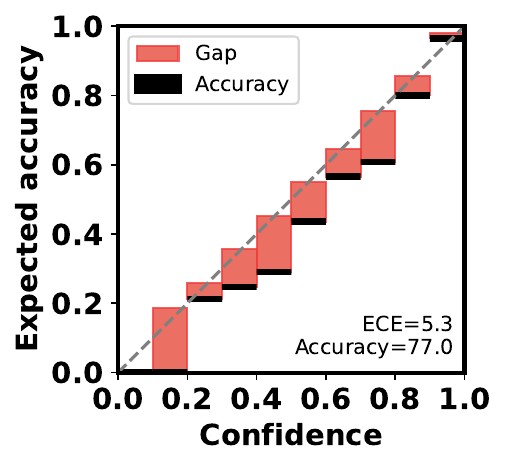}
        \caption{Flowers \ours{}}
    \end{subfigure}
    \begin{subfigure}[b]{0.32\linewidth}
        \centering
        \includegraphics[width=\linewidth]{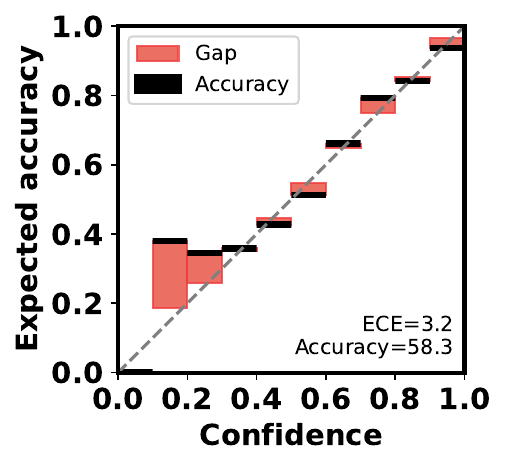}
        \caption{EuroSAT \ours{}}
    \end{subfigure}
    \begin{subfigure}[b]{0.32\linewidth}
        \centering
        \includegraphics[width=\linewidth]{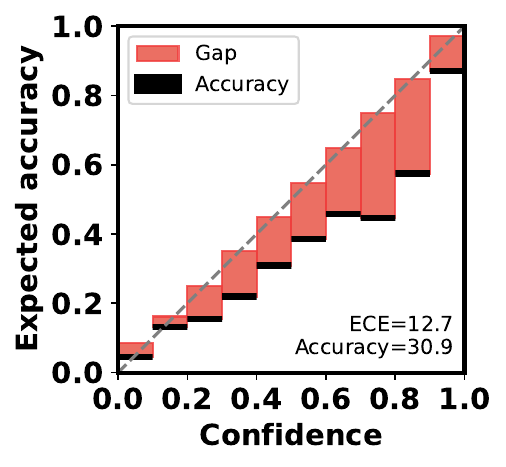}
        \caption{Aircraft \ours{}}
    \end{subfigure}
    
    \caption{\textbf{Reliability diagrams of O-TPT vs \ours{}.} Plots showing the calibration error across the Flowers102, EuroSAT and FGVC Aircraft datasets for O-TPT (top row) and \ours{} (bottom row).}
    \label{fig:reliability}
\end{figure}

In addition to these numerical results, we depict reliability diagrams (Fig.~\ref{fig:reliability}) across three representative datasets, which can be used to show not only the extent of miscalibration, but also the direction of the miscalibration, i.e., whether the model is systematically overconfident or underconfident. Looking closely at these plots, they reveal that O-TPT exhibits pronounced overconfidence, with predicted probabilities consistently exceeding observed accuracies. In contrast, \ours{} yields flatter reliability curves that more closely follow the diagonal, indicating better alignment between confidence and accuracy. This reduction directly translates into lower ECE scores, and supports the theoretical findings presented in Corollary~\ref{cor:confidence_increase}, which show that full orthogonality increases the confidence floor more aggressively than our Huber-like regularizer, virtually increasing confidence.

\noindent \textbf{Robustness under natural distributional drifts.} Tab.~\ref{tab:vitl_shift} presents the results across the four different variants of ImageNet, which present a distributional drift. In particular, we can observe that while performing on par with O-TPT in terms of accuracy, it decreases the ECE by 1.5. Similarly, even though TPT and C-TPT yield higher accuracy, calibration metrics are substantially improved, e.g., 5.8 and 4.2, compared to TPT and C-TPT, respectively. These results align with the observations in O-TPT, where TPT and C-TPT achieved better discriminative performance, at the cost of large degradations in calibration. Together, these outcomes underscore the ability of our approach to adapt to natural distribution shifts while enhancing the uncertainty. 
\begin{table}[h!]
    \caption{\textbf{Distribution shift (ViT-L/14 backbone).} Accuracy and ECE across four variants of the ImageNet dataset compared to the four compared baselines. Green and red indicate positive and negative changes with respect to O-TPT, respectively.}
    \label{tab:vitl_shift}
    \centering
    \resizebox{\linewidth}{!}{
    \begin{tabular}{l l c c c c c}
        \toprule
         & & -A
         & -v2
         & -R
         & -Sketch
         & Average \\
        \midrule
        \multirow{5}{*}{\rotatebox{90}{\textbf{Accuracy}}} & Zero-Shot &
            68.8 &
            67.9 &
            85.4 &
            57.8 &
            70.0 \\
        & TPT &
            73.9 &
            70.0 &
            87.8 &
            59.7 &
            72.9 \\
        & C-TPT &
            72.0 &
            69.9 &
            86.7 &
            59.5 &
            72.0 \\
        & O-TPT &
            71.3 &
            69.0 &
            86.0 &
            58.8 &
            71.3 \\
        & \our \textbf{\ours{} (\textit{Ours})} &
            \our 70.4\wor{-0.9} &
            \our 68.9\wor{-0.1} &
            \our 86.8\imp{+0.9} &
            \our 58.9\imp{+0.1} &
            \our 71.3\imp{+0.0} \\
        \midrule
        \multirow{5}{*}{\rotatebox{90}{\textbf{ECE}}} & Zero-Shot &
            3.9 &
            3.9 &
            2.5 &
            7.4 &
            4.4 \\
        & TPT &
            13.1 &
            16.4 &
            5.0 &
            22.2 &
            14.2 \\
        & C-TPT &
            11.7 &
            14.6 &
            3.5 &
            20.4 &
            12.6 \\
        & O-TPT &
            10.8 &
            10.6 &
            2.7 &
            15.3 &
            9.9 \\
        & \our \textbf{\ours{} (\textit{Ours})} &
            \our 7.3\imp{-3.5} &
            \our 10.4\imp{-0.2} &
            \our 1.4\imp{-1.3} &
            \our 14.4\imp{-0.9} &
            \our 8.4\imp{-1.5} \\
        \bottomrule
    \end{tabular}
    }
\end{table}

\begin{figure}[h]
    \centering
    \begin{subfigure}[b]{0.49\linewidth}
        \centering
        \includegraphics[width=\linewidth]{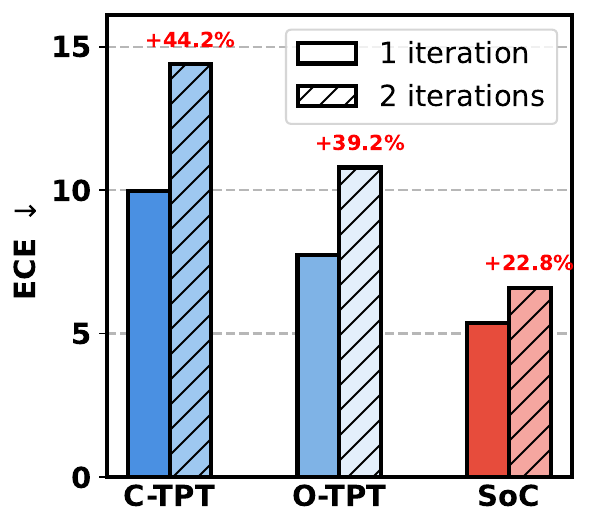}
    \end{subfigure}
    \begin{subfigure}[b]{0.49\linewidth}
        \centering
        \includegraphics[width=\linewidth]{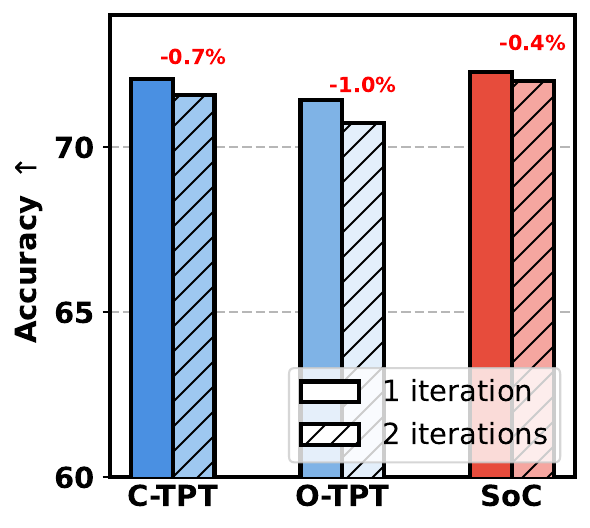}
    \end{subfigure}
    \caption{\textbf{ECE and accuracy for one and two gradient steps.} Standard one-step and two-step gradient updates for C-TPT, O-TPT, and \ours{} averaged across 11 datasets. Hatched bars indicate the result when applying two gradient updates over text prompts.}
    \label{fig:2steps}
\end{figure}

\noindent \textbf{O-TPT confidence significantly increases in multi-step TPT.}  TPT-based approaches typically perform a single update of the prompts. Under this well-established scenario, and motivated by our analytical findings on confidence inflation under full orthogonality (\textit{Proposition~\ref{prop:bound}}, \textit{Corollary~\ref{cor:confidence_increase}}), we empirically investigate the impact of applying two gradient updates on calibration and performance, whose results are shown in Fig.~\ref{fig:2steps}. While C-TPT, O-TPT, and \ours{} were all negatively impacted by performing an additional gradient update, the effect was much less pronounced on our method, particularly from a calibration standpoint. Indeed, while the ECE obtained by \ours{} degraded nearly 23\% after a second gradient update, O-TPT calibration deteriorated by 39\%, nearly twice as much. This behavior aligns with our first-order analysis (Section \ref{subsec:first_order}), which shows that full orthogonality induces sharper reductions in worst-case similarity $\mu$ per gradient step, leading to virtually increased confidence, even when unnecessary. 

\begin{table}[b]
    \caption{\textbf{Robustness to backbone.} Average accuracy and ECE across all 11 datasets for two sizes of 
    ViTs. Differences with respect to O-TPT are highlighted in green.}
    \label{tab:model_agnostic}
    \centering
    \footnotesize
    \begin{tabular}{l c c c c}
    \toprule
     & \multicolumn{2}{c}{ViT-B/16} & \multicolumn{2}{c}{ViT-L/14} \\
    \cmidrule(lr){2-3} \cmidrule(lr){4-5}
    &  Acc. & ECE & Acc. & ECE \\ 
    \midrule
          Zero-Shot & 63.9 & 4.2 & 71.1 & 5.1 \\
          TPT & 65.2 & 11.2 & 72.0 & 14.9 \\
          C-TPT & 64.6 & 5.0 & 72.1 & 10.0 \\
          O-TPT & 64.0 & 4.8 & 71.4 & 7.7 \\
          \our\textbf{\ours{} (\textit{Ours})} &
            \our64.6\imp{+0.6} &
            \our4.3\imp{-0.5} &
            \our72.3\imp{+0.9} &
            \our5.4\imp{-2.3} \\
    \bottomrule
    \end{tabular}
\end{table}
\mypar{\textbf{Robustness to backbone}}. To assess the robustness of our method across different backbone architectures, we extend our evaluation and include results using ViT-B/16. This comparison allows us to verify whether the calibration and alignment benefits inherent in our regularizer persist under smaller, less expressive models (Tab.~\ref{tab:model_agnostic}). In particular, we observe a similar trend to that observed in the ViT-L/14 backbone, i.e., \ours{} outperforms state-of-the-art O-TPT in accuracy and ECE. Furthermore, it is interesting to note that, while \ours{} enhances the discriminative performance of zero-shot (ZS) predictions (+0.7 and +1.2), it maintains highly comparable ECE scores across backbones (4.3 \textit{vs} 4.2; 5.4 \textit{vs} 5.1). This highlights the strong calibration properties of our approach, alongside with its ability to improve ZS performance.

\begin{figure}[h!]
    \centering
    \begin{subfigure}[b]{0.49\linewidth}
        \centering
        \includegraphics[width=\linewidth]{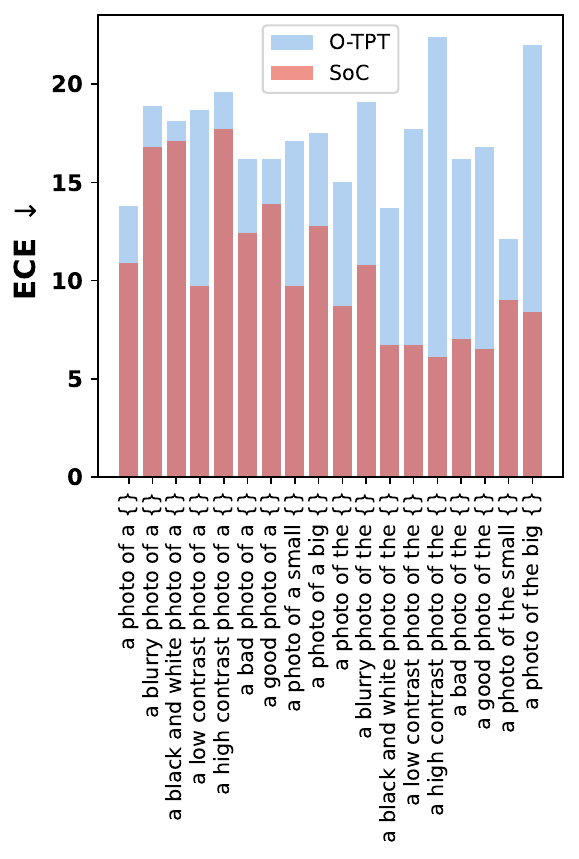}
    \end{subfigure}
    \begin{subfigure}[b]{0.49\linewidth}
        \centering
        \includegraphics[width=\linewidth]{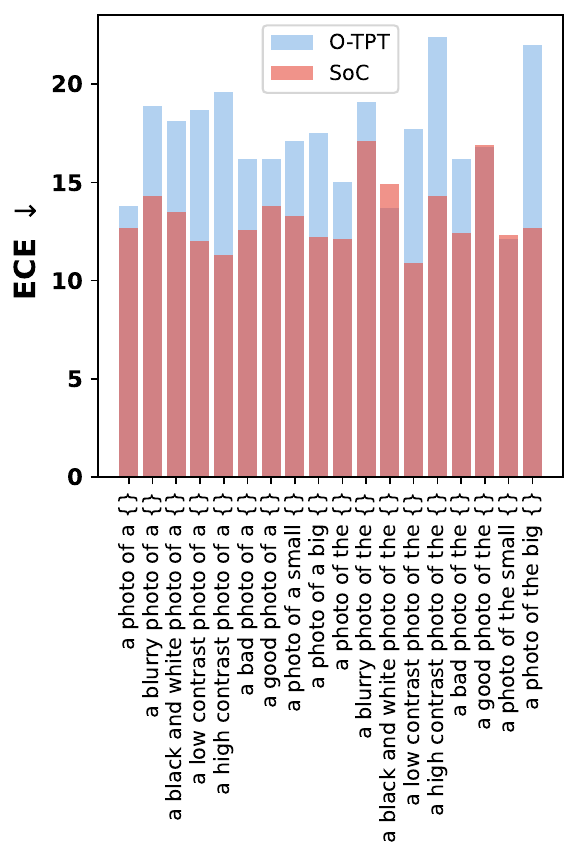}
    \end{subfigure}
    \caption{\textbf{Calibration sensitivity of O-TPT vs. \ours{} for various initial text prompts.} ECE on the DTD (\textit{left}) and Aircraft (\textit{right}) datasets for 18 different prompts from CLIP~\cite{radford2021clip}.}
    \label{fig:varying_prompts}
\end{figure}

\noindent \textbf{Calibration across prompt variability.} Prompt initialization remains a critical factor in CLIP-based adaptation, as different textual templates can induce large variations in calibration. Fig.~\ref{fig:varying_prompts} depicts ECE across 18 CLIP text prompts for two different datasets. While both O-TPT and \ours{} exhibit sensitivity, an interesting observation is that our approach yields lower ECE across nearly all prompts. In other words, even though calibration still depends on prompt initialization, our regularizer typically limits the extreme overconfidence spikes characteristic of O-TPT. These results underscore that \ours{} delivers more reliable predictions than O-TPT, regardless of the chosen template. 

\noindent \textbf{Prompt initialization with CoOp.} CoOp~\cite{zhou2022coop} is a prompt tuning method for VLMs, which can be used to optimize the initial prompt in a few-shot setting. As in O-TPT~\cite{sharifdeen2025tpt}, we evaluate the prompt embeddings trained in a supervised manner, using $k$-shots, to evaluate the calibration effectiveness of these prompts with \ours{} during test-time prompt tuning. Tab.~\ref{tab:coop} shows results when using CoOp-initialized prompts, trained for 2-shots and 4-shots. \ours{} keeps outperforming O-TPT both in terms of accuracy and calibration, even though the gap get smaller as the accuracy increases and the calibration error decreases.\\

\begin{table}[h!]
    \caption{\textbf{Effect of pretraining the prompts with CoOp.} Results show the average across all 11 datasets for 2-shot and 4-shot pretrained prompts using CoOp. Baseline represents using the non-CoOp-initialized prompts. Differences with respect to O-TPT are highlighted in green.}
    \label{tab:coop}
    \centering
    \resizebox{0.8\linewidth}{!}{
    \begin{tabular}{l l c c c}
    \toprule
    
    & & Baseline & 2-shots & 4-shots \\ 
    \midrule
         \multirow{4}{*}{\rotatebox{90}{\textbf{Accuracy}}}
         & TPT$_\text{+CoOp}$ & 72.0 & 76.0 & 80.2 \\
         & C-TPT$_\text{+CoOp}$ & 72.1 & 74.1 & 78.3 \\
         & O-TPT$_\text{+CoOp}$ & 71.4 & 72.9 & 76.6 \\
         & \our\textbf{\ours{}$_\text{+CoOp}$ (\textit{Ours})} &
            \our72.3\imp{+0.9} &
            \our75.2\imp{+2.3}&
            \our78.0\imp{+1.4} \\
    \midrule
         \multirow{4}{*}{\rotatebox{90}{\textbf{ECE}}}
         & TPT$_\text{+CoOp}$ & 14.9 & 14.8 & 11.7 \\
         & C-TPT$_\text{+CoOp}$ & 10.0 & 10.4 & 7.7\\
         & O-TPT$_\text{+CoOp}$ & 7.7 & 7.4 & 5.6\\
         & \our \textbf{\ours{}$_\text{+CoOp}$ (\textit{Ours})} &
            \our5.4\imp{-2.3} &
            \our6.3\imp{-1.1} &
            \our5.4\imp{-0.2} \\
    \bottomrule
    \end{tabular}
    }
\end{table}

\begin{figure}
    \centering
    \includegraphics[width=0.8\linewidth]{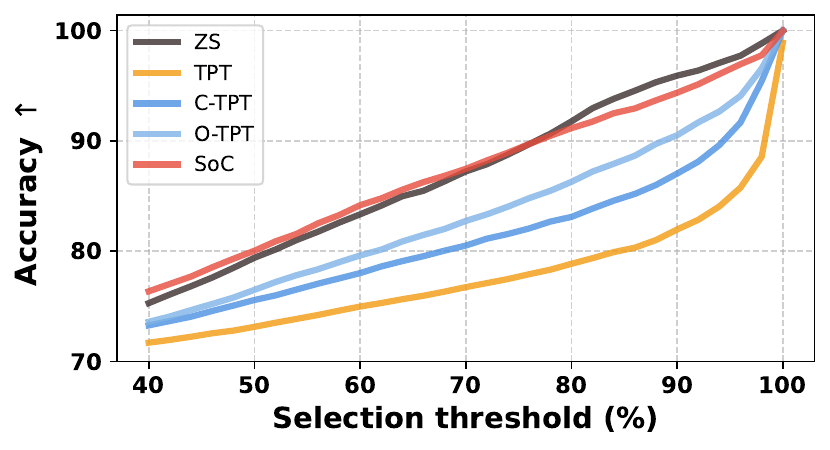}
    \caption{\textbf{Selective accuracy across different thresholds}. For each threshold, we select only the samples whose confidence (i.e., maximum of softmax) exceeds that value, and compute the accuracy on this subset.}
    \label{fig:sel-acc}
\end{figure}

\noindent \textbf{When the model is confident, how often is it correct?} To further evaluate the reliability of our method, we report selective classification accuracy under varying confidence thresholds (Fig.~\ref{fig:sel-acc}). This setting reflects practical deployment scenarios where predictions are accepted only if the model’s confidence exceeds a predefined threshold, making calibration critical for safe decision-making. Across all thresholds, \ours{} consistently outperforms TPT, C-TPT, and O-TPT, achieving higher selective accuracy, with gaps often ranging between 5-10\%. Notably, across all thresholds, our method matches the performance of the ZS baseline, despite being adapted on unlabeled test data. This indicates that our regularizer preserves semantic alignment while improving calibration. It is important to note that, even though ZS CLIP appears strong in this setting, it benefits from fixed, pretrained text embeddings that are not exposed to entropy minimization or adaptation-induced drift. However, as shown in prior experiments, it lacks flexibility and cannot adapt to domain-specific semantics. In contrast, our method retains the robustness of ZS while enabling task-specific adaptation, yielding superior performance at lower thresholds, where calibration plays a larger role in filtering uncertain predictions.

These results reinforce our core claim: \textit{by limiting confidence increase and preserving semantic proximity, \ours{} enables reliable adaptation without sacrificing reliability}. This makes it especially suitable for real-world applications, where selective prediction is paramount.

\begin{figure}[h]
    \centering
    \includegraphics[width=\linewidth]{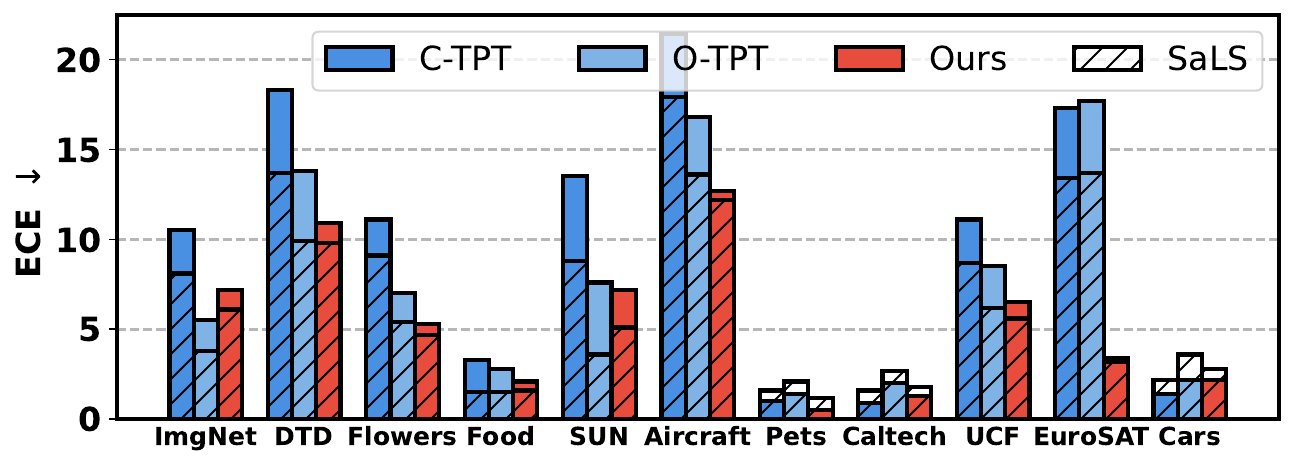}
    \caption{\textbf{ECE with and without SaLS.} ECE for C-TPT, O-TPT, and \ours{} with and without applying SaLS for further calibration.}
    \label{fig:sals}
\end{figure}

\noindent \textbf{Further improving \ours{} performance.} SaLS \cite{murugesan2024robust} presented a simple, model-agnostic post-hoc strategy to enhance calibration of TPT, among other adaptation strategies. To assess the compatibility of our method with post-hoc calibration strategies, we apply SaLS to C-TPT, O-TPT, and \ours{}. These results, depicted in Fig.~\ref{fig:sals}, demonstrate that SaLS often enhances the calibration results provided by our approach, showcasing its complementarity. Indeed, even when SaLS is applied across all methods, our approach remains the best calibrated across nearly all datasets. Last, it is noteworthy to highlight that the improvements from SaLS are notably smaller on our method, which indicates that the predictions from \ours{} are often already well-calibrated prior to post-hoc adjustment, demonstrating the stronger calibration capabilities of the proposed approach.

Additional experiments and results are presented in Appendix Section~\ref{sec:additional_experiments}.

\section{Conclusion}
\label{sec:conclusion}

In this work, we have shown that enforcing full orthogonality (i.e., O-TPT) in test-time prompt tuning, while intuitively appealing, systematically distorts semantic structure and inflates confidence, potentially amplifying miscalibration. This issue is particularly magnified in classes with high semantic similarity, which are naturally close in both image and text embedding manifolds. To address this limitation, we have presented \textbf{\ours{}}, which replaces the strong repulsion in O-TPT with a Huber-based regularizer that respects semantic proximity, yielding smoother prototype geometry and improved calibration. To support our theoretical findings, comprehensive experiments across diverse datasets and backbones demonstrate that \ours{} consistently outperforms prior calibration-oriented TPT methods, while preserving competitive accuracy. We believe that our analysis on the impact of enforcing full orthogonality for highly similar pairs opens the door to more semantically-aware calibration methods for VLMs.

\mypar{Acknowledgments.} This work has benefited from state financial aid, managed by the Agence Nationale de Recherche under the investment program integrated into France 2030, project reference ANR-21-RHUS-0003. We acknowledge the support of the Natural Sciences and Engineering Research Council of Canada (NSERC). This work was granted access to the HPC resources of IDRIS under the allocation 2024-AD011014802R1 made by GENCI. The author gratefully acknowledges support from ILLS during the internship in which this work was conducted.
{
    \small
    \bibliographystyle{ieeenat_fullname}
    \bibliography{main}

\begin{thebibliography}{56}
\providecommand{\natexlab}[1]{#1}
\providecommand{\url}[1]{\texttt{#1}}
\expandafter\ifx\csname urlstyle\endcsname\relax
  \providecommand{\doi}[1]{doi: #1}\else
  \providecommand{\doi}{doi: \begingroup \urlstyle{rm}\Url}\fi

\bibitem[Alayrac et~al.(2022)Alayrac, Donahue, Luc, Miech, Barr, Hasson, Lenc, Mensch, Millicah, Reynolds, Ring, Rutherford, Cabi, Han, Gong, Samangooei, Monteiro, Menick, Borgeaud, Brock, Nematzadeh, Sharifzadeh, Binkowski, Barreira, Vinyals, Zisserman, and Simonyan]{alayrac2022flamingo}
Jean-Baptiste Alayrac, Jeff Donahue, Pauline Luc, Antoine Miech, Iain Barr, Yana Hasson, Karel Lenc, Arthur Mensch, Katie Millicah, Malcolm Reynolds, Roman Ring, Eliza Rutherford, Serkan Cabi, Tengda Han, Zhitao Gong, Sina Samangooei, Marianne Monteiro, Jacob Menick, Sebastian Borgeaud, Andrew Brock, Aida Nematzadeh, Sahand Sharifzadeh, Mikolaj Binkowski, Ricardo Barreira, Oriol Vinyals, Andrew Zisserman, and Karen Simonyan.
\newblock Flamingo: a visual language model for few-shot learning.
\newblock In \emph{Proceedings of the 36th International Conference on Neural Information Processing Systems}, Red Hook, NY, USA, 2022. Curran Associates Inc.

\bibitem[Bossard et~al.(2014)Bossard, Guillaumin, and Van~Gool]{bossard14}
Lukas Bossard, Matthieu Guillaumin, and Luc Van~Gool.
\newblock Food-101 -- mining discriminative components with random forests.
\newblock In \emph{European Conference on Computer Vision}, 2014.

\bibitem[Cheng and Vasconcelos(2022)]{cheng2022calibrating}
Jiacheng Cheng and Nuno Vasconcelos.
\newblock Calibrating deep neural networks by pairwise constraints.
\newblock In \emph{Proceedings of the IEEE/CVF Conference on Computer Vision and Pattern Recognition}, pages 13709--13718, 2022.

\bibitem[Cimpoi et~al.(2014)Cimpoi, Maji, Kokkinos, Mohamed, , and Vedaldi]{cimpoi14describing}
M. Cimpoi, S. Maji, I. Kokkinos, S. Mohamed, , and A. Vedaldi.
\newblock Describing textures in the wild.
\newblock In \emph{IEEE Conference on Computer Vision and Pattern Recognition}, 2014.

\bibitem[Deng et~al.(2009)Deng, Dong, Socher, Li, Li, and Fei-Fei]{5206848}
Jia Deng, Wei Dong, Richard Socher, Li-Jia Li, Kai Li, and Li Fei-Fei.
\newblock Imagenet: A large-scale hierarchical image database.
\newblock In \emph{2009 IEEE Conference on Computer Vision and Pattern Recognition}, pages 248--255, 2009.

\bibitem[Guo et~al.(2017)Guo, Pleiss, Sun, and Weinberger]{guo2017calibration}
Chuan Guo, Geoff Pleiss, Yu Sun, and Kilian~Q. Weinberger.
\newblock On calibration of modern neural networks.
\newblock In \emph{Proceedings of the 34th International Conference on Machine Learning - Volume 70}, page 1321–1330. JMLR.org, 2017.

\bibitem[Hantao~Yao(2023)]{kgcoop23}
Changsheng~Xu Hantao~Yao, Rui~Zhang.
\newblock Visual-language prompt tuning with knowledge-guided context optimization.
\newblock In \emph{The IEEE/CVF Conference on Computer Vision and Pattern Recognition}, 2023.

\bibitem[Helber et~al.(2019)Helber, Bischke, Dengel, and Borth]{helber2017eurosat}
Patrick Helber, Benjamin Bischke, Andreas Dengel, and Damian Borth.
\newblock Eurosat: A novel dataset and deep learning benchmark for land use and land cover classification.
\newblock \emph{IEEE Journal of Selected Topics in Applied Earth Observations and Remote Sensing}, 2019.

\bibitem[Hendrycks et~al.(2020)Hendrycks, Mu, Cubuk, Zoph, Gilmer, and Lakshminarayanan]{hendrycks2020augmix}
Dan Hendrycks, Norman Mu, Ekin~D. Cubuk, Barret Zoph, Justin Gilmer, and Balaji Lakshminarayanan.
\newblock Augmix: A simple data processing method to improve robustness and uncertainty.
\newblock In \emph{International Conference on Learning Representations (ICLR)}, 2020.

\bibitem[Hendrycks et~al.(2021{\natexlab{a}})Hendrycks, Basart, Mu, Kadavath, Wang, Dorundo, Desai, Zhu, Parajuli, Guo, Song, Steinhardt, and Gilmer]{hendrycks2021many}
Dan Hendrycks, Steven Basart, Norman Mu, Saurav Kadavath, Frank Wang, Evan Dorundo, Rahul Desai, Tyler Zhu, Samyak Parajuli, Mike Guo, Dawn Song, Jacob Steinhardt, and Justin Gilmer.
\newblock The many faces of robustness: A critical analysis of out-of-distribution generalization.
\newblock \emph{ICCV}, 2021{\natexlab{a}}.

\bibitem[Hendrycks et~al.(2021{\natexlab{b}})Hendrycks, Zhao, Basart, Steinhardt, and Song]{imagenet_a}
Dan Hendrycks, Kevin Zhao, Steven Basart, Jacob Steinhardt, and Dawn Song.
\newblock Natural adversarial examples.
\newblock In \emph{Proceedings of the IEEE/CVF Conference on Computer Vision and Pattern Recognition (CVPR)}, pages 15262--15271, 2021{\natexlab{b}}.

\bibitem[Huber(1964)]{huber1964robust}
Peter~J. Huber.
\newblock Robust estimation of a location parameter.
\newblock \emph{Annals of Mathematical Statistics}, 35\penalty0 (1):\penalty0 73--101, 1964.

\bibitem[Jia et~al.(2021)Jia, Yang, Xia, Chen, Parekh, Pham, Le, Sung, Li, and Duerig]{jia2021scaling}
Chao Jia, Yinfei Yang, Ye Xia, Yi-Ting Chen, Zarana Parekh, Hieu Pham, Quoc~V. Le, Yun-Hsuan Sung, Zhen Li, and Tom Duerig.
\newblock Scaling up visual and vision-language representation learning with noisy text supervision.
\newblock In \emph{International Conference on Machine Learning}, 2021.

\bibitem[Khandelwal et~al.(2022)Khandelwal, Weihs, Mottaghi, and Kembhavi]{khandelwal2022simple}
Apoorv Khandelwal, Luca Weihs, Roozbeh Mottaghi, and Aniruddha Kembhavi.
\newblock Simple but effective: {CLIP} embeddings for embodied {AI}.
\newblock In \emph{Proceedings of the IEEE/CVF Conference on Computer Vision and Pattern Recognition}, pages 14829--14838, 2022.

\bibitem[Koleilat et~al.(2025)Koleilat, Asgariandehkordi, Rivaz, and Xiao]{koleilat2025biomedcoop}
Taha Koleilat, Hojat Asgariandehkordi, Hassan Rivaz, and Yiming Xiao.
\newblock Biomedcoop: Learning to prompt for biomedical vision-language models.
\newblock In \emph{Proceedings of the Computer Vision and Pattern Recognition Conference}, pages 14766--14776, 2025.

\bibitem[Krause et~al.(2013)Krause, Stark, Deng, and Fei-Fei]{KrauseStarkDengFei-Fei_3DRR2013}
Jonathan Krause, Michael Stark, Jia Deng, and Li Fei-Fei.
\newblock 3d object representations for fine-grained categorization.
\newblock In \emph{4th International IEEE Workshop on 3D Representation and Recognition (3dRR-13)}, Sydney, Australia, 2013.

\bibitem[Li et~al.(2006)Li, Fergus, and Perona]{li2006one}
Fei-Fei Li, Rob Fergus, and Pietro Perona.
\newblock One-shot learning of object categories.
\newblock \emph{IEEE Transactions on Pattern Analysis and Machine Intelligence}, 2006.

\bibitem[Li et~al.(2021)Li, Selvaraju, Gotmare, Joty, Xiong, and Hoi]{ALBEF}
Junnan Li, Ramprasaath~R. Selvaraju, Akhilesh~Deepak Gotmare, Shafiq Joty, Caiming Xiong, and Steven Hoi.
\newblock Align before fuse: Vision and language representation learning with momentum distillation.
\newblock In \emph{NeurIPS}, 2021.

\bibitem[Li et~al.(2022)Li, Li, Xiong, and Hoi]{li2022blip}
Junnan Li, Dongxu Li, Caiming Xiong, and Steven C.~H. Hoi.
\newblock Blip: Bootstrapping language-image pre-training for unified vision-language understanding and generation.
\newblock In \emph{International Conference on Machine Learning}, 2022.

\bibitem[Li et~al.(2023)Li, Sun, and Li]{li2023clip}
Siyuan Li, Li Sun, and Qingli Li.
\newblock {CLIP}-re{ID}: exploiting vision-language model for image re-identification without concrete text labels.
\newblock In \emph{Proceedings of the AAAI conference on artificial intelligence}, pages 1405--1413, 2023.

\bibitem[Liu et~al.(2022)Liu, Ben~Ayed, Galdran, and Dolz]{liu2022devil}
Bingyuan Liu, Ismail Ben~Ayed, Adrian Galdran, and Jose Dolz.
\newblock The devil is in the margin: Margin-based label smoothing for network calibration.
\newblock In \emph{Proceedings of the IEEE/CVF Conference on Computer Vision and Pattern Recognition (CVPR)}, pages 80--88, 2022.

\bibitem[Liu et~al.(2023{\natexlab{a}})Liu, Rony, Galdran, Dolz, and Ben~Ayed]{liu2023class}
Bingyuan Liu, J{\'e}r{\^o}me Rony, Adrian Galdran, Jose Dolz, and Ismail Ben~Ayed.
\newblock Class adaptive network calibration.
\newblock In \emph{Conference on Computer Vision and Pattern Recognition (CVPR)}, pages 16070--16079, 2023{\natexlab{a}}.

\bibitem[Liu et~al.(2023{\natexlab{b}})Liu, Li, Wu, and Lee]{liu2023llava}
Haotian Liu, Chunyuan Li, Qingyang Wu, and Yong~Jae Lee.
\newblock Visual instruction tuning, 2023{\natexlab{b}}.

\bibitem[Loshchilov and Hutter(2019)]{loshchilov2019decoupled}
Ilya Loshchilov and Frank Hutter.
\newblock Decoupled weight decay regularization.
\newblock In \emph{International Conference on Learning Representations (ICLR)}, 2019.

\bibitem[Lv et~al.(2025)Lv, Chen, Zhou, Li, and Guo]{lv2025contrast}
Song-Lin Lv, Yu-Yang Chen, Zhi Zhou, Yu-Feng Li, and Lan-Zhe Guo.
\newblock Contrast-aware calibration for fine-tuned {CLIP}: Leveraging image-text alignment.
\newblock \emph{arXiv preprint arXiv:2501.19060}, 2025.

\bibitem[Maji et~al.(2013)Maji, Kannala, Rahtu, Blaschko, and Vedaldi]{maji13fine-grained}
S. Maji, J. Kannala, E. Rahtu, M. Blaschko, and A. Vedaldi.
\newblock Fine-grained visual classification of aircraft.
\newblock Technical report, 2013.

\bibitem[Manli et~al.(2022)Manli, Weili, De-An, Zhiding, Tom, Anima, and Chaowei]{shu2022tpt}
Shu Manli, Nie Weili, Huang De-An, Yu Zhiding, Goldstein Tom, Anandkumar Anima, and Xiao Chaowei.
\newblock Test-time prompt tuning for zero-shot generalization in vision-language models.
\newblock In \emph{NeurIPS}, 2022.

\bibitem[Minderer et~al.(2021)Minderer, Djolonga, Romijnders, Hubis, Zhai, Houlsby, Tran, and Lucic]{NEURIPS2021_8420d359}
Matthias Minderer, Josip Djolonga, Rob Romijnders, Frances Hubis, Xiaohua Zhai, Neil Houlsby, Dustin Tran, and Mario Lucic.
\newblock Revisiting the calibration of modern neural networks.
\newblock In \emph{Advances in Neural Information Processing Systems}, pages 15682--15694. Curran Associates, Inc., 2021.

\bibitem[Morales-{\'A}lvarez et~al.(2025)Morales-{\'A}lvarez, Christodoulidis, Vakalopoulou, Piantanida, and Dolz]{morales2025bayesadapter}
Pablo Morales-{\'A}lvarez, Stergios Christodoulidis, Maria Vakalopoulou, Pablo Piantanida, and Jose Dolz.
\newblock Bayesadapter: enhanced uncertainty estimation in {CLIP} few-shot adaptation.
\newblock \emph{International Journal of Computer Vision (IJCV)}, 2025.

\bibitem[Mu et~al.(2022)Mu, Kirillov, Wagner, and Xie]{mu2022slip}
Norman Mu, Alexander Kirillov, David Wagner, and Saining Xie.
\newblock Slip: Self-supervision meets language-image pre-training.
\newblock In \emph{Computer Vision -- ECCV 2022}, pages 514--532. Springer International Publishing, 2022.

\bibitem[Mukhoti et~al.(2020)]{mukhoti2020calibratingSup}
Jishnu Mukhoti et~al.
\newblock Calibrating deep neural networks using focal loss.
\newblock In \emph{Advances in Neural Information Processing Systems (NeurIPS)}, 2020.

\bibitem[M{\"u}ller et~al.(2019)]{muller2019does}
Rafael M{\"u}ller et~al.
\newblock When does label smoothing help?
\newblock \emph{NeurIPS}, 32, 2019.

\bibitem[Murugesan et~al.(2024)Murugesan, Silva-Rodriguez, Ayed, and Dolz]{murugesan2024robust}
Balamurali Murugesan, Julio Silva-Rodriguez, Ismail~Ben Ayed, and Jose Dolz.
\newblock Robust calibration of large vision-language adapters.
\newblock In \emph{European Conference on Computer Vision (ECCV)}, 2024.

\bibitem[Nilsback and Zisserman(2008)]{Nilsback08}
M-E. Nilsback and A. Zisserman.
\newblock Automated flower classification over a large number of classes.
\newblock In \emph{Indian Conference on Computer Vision, Graphics and Image Processing}, 2008.

\bibitem[Oh et~al.(2024)Oh, Lim, Kim, Han, Yun, Choo, Hauptmann, Cheng, and Song]{oh2024towards}
Changdae Oh, Hyesu Lim, Mijoo Kim, Dongyoon Han, Sangdoo Yun, Jaegul Choo, Alexander Hauptmann, Zhi-Qi Cheng, and Kyungwoo Song.
\newblock Towards calibrated robust fine-tuning of vision-language models.
\newblock \emph{Advances in Neural Information Processing Systems}, 37:\penalty0 12677--12707, 2024.

\bibitem[Pan et~al.(2024)Pan, Yaman, Nesti, Mallik, Allievi, Velipasalar, and Ren]{pan2024vlp}
Chenbin Pan, Burhaneddin Yaman, Tommaso Nesti, Abhirup Mallik, Alessandro~G Allievi, Senem Velipasalar, and Liu Ren.
\newblock {VLP}: Vision language planning for autonomous driving.
\newblock In \emph{Proceedings of the IEEE/CVF Conference on Computer Vision and Pattern Recognition}, pages 14760--14769, 2024.

\bibitem[Parkhi et~al.(2012)Parkhi, Vedaldi, Zisserman, and Jawahar]{parkhi12a}
O.~M. Parkhi, A. Vedaldi, A. Zisserman, and C.~V. Jawahar.
\newblock Cats and dogs.
\newblock In \emph{IEEE Conference on Computer Vision and Pattern Recognition}, 2012.

\bibitem[Pereyra et~al.(2017)]{pereyra2017regularizing}
Gabriel Pereyra et~al.
\newblock Regularizing neural networks by penalizing confident output distributions.
\newblock In \emph{International Conference on Learning Representations (ICLR)}, 2017.

\bibitem[Radford et~al.(2021)Radford, Kim, Hallacy, Ramesh, Goh, Agarwal, Sastry, Askell, Mishkin, Clark, Krueger, and Sutskever]{radford2021clip}
Alec Radford, Jong~Wook Kim, Chris Hallacy, Aditya Ramesh, Gabriel Goh, Sandhini Agarwal, Girish Sastry, Amanda Askell, Pamela Mishkin, Jack Clark, Gretchen Krueger, and Ilya Sutskever.
\newblock Learning transferable visual models from natural language supervision.
\newblock In \emph{International Conference on Machine Learning}, 2021.

\bibitem[Recht et~al.(2019)Recht, Roelofs, Schmidt, and Shankar]{pmlr-v97-recht19a}
Benjamin Recht, Rebecca Roelofs, Ludwig Schmidt, and Vaishaal Shankar.
\newblock Do {I}mage{N}et classifiers generalize to {I}mage{N}et?
\newblock In \emph{Proceedings of the 36th International Conference on Machine Learning}, pages 5389--5400. PMLR, 2019.

\bibitem[Sharifdeen et~al.(2025)Sharifdeen, Munir, Baliah, Khan, and Khan]{sharifdeen2025tpt}
Ashshak Sharifdeen, Muhammad~Akhtar Munir, Sanoojan Baliah, Salman Khan, and Muhammad~Haris Khan.
\newblock O-{TPT}: Orthogonality constraints for calibrating test-time prompt tuning in vision-language models.
\newblock In \emph{Proceedings of the Computer Vision and Pattern Recognition Conference}, pages 19942--19951, 2025.

\bibitem[Shu et~al.(2022)Shu, Nie, Huang, Yu, Goldstein, Anandkumar, and Xiao]{shu2022test}
Manli Shu, Weili Nie, De-An Huang, Zhiding Yu, Tom Goldstein, Anima Anandkumar, and Chaowei Xiao.
\newblock Test-time prompt tuning for zero-shot generalization in vision-language models.
\newblock \emph{Advances in Neural Information Processing Systems}, 35:\penalty0 14274--14289, 2022.

\bibitem[Silva-Rodr{\'\i}guez et~al.(2025)Silva-Rodr{\'\i}guez, Shakeri, Bahig, Dolz, and Ben~Ayed]{silva2025few}
Julio Silva-Rodr{\'\i}guez, Fereshteh Shakeri, Houda Bahig, Jose Dolz, and Ismail Ben~Ayed.
\newblock Few-shot, now for real: Medical {VLM}s adaptation without balanced sets or validation.
\newblock In \emph{International Conference on Medical Image Computing and Computer-Assisted Intervention}, pages 237--247. Springer, 2025.

\bibitem[Soomro et~al.(2012)Soomro, Zamir, and Shah]{Soomro2012UCF101AD}
Khurram Soomro, Amir Zamir, and Mubarak Shah.
\newblock Ucf101: A dataset of 101 human actions classes from videos in the wild.
\newblock \emph{ArXiv}, abs/1212.0402, 2012.

\bibitem[Tu et~al.(2024)Tu, Deng, Campbell, Gould, and Gedeon]{tu2024empirical}
Weijie Tu, Weijian Deng, Dylan Campbell, Stephen Gould, and Tom Gedeon.
\newblock An empirical study into what matters for calibrating vision-language models.
\newblock In \emph{International Conference on Machine Learning}, pages 48791--48808. PMLR, 2024.

\bibitem[Wang et~al.(2019)Wang, Ge, Lipton, and Xing]{wang2019learning}
Haohan Wang, Songwei Ge, Zachary Lipton, and Eric~P Xing.
\newblock Learning robust global representations by penalizing local predictive power.
\newblock In \emph{Advances in Neural Information Processing Systems}, pages 10506--10518, 2019.

\bibitem[Wang et~al.(2024)Wang, Wang, Wang, Zhang, Zhou, and Wei]{wang2024open}
Shuoyuan Wang, Jindong Wang, Guoqing Wang, Bob Zhang, Kaiyang Zhou, and Hongxin Wei.
\newblock Open-vocabulary calibration for fine-tuned {CLIP}.
\newblock In \emph{International Conference on Machine Learning}, pages 51734--51754. PMLR, 2024.

\bibitem[Xiao et~al.(2010)Xiao, Hays, Ehinger, Oliva, and Torralba]{xiao2010sun}
Jianxiong Xiao, James Hays, Krista~A Ehinger, Aude Oliva, and Antonio Torralba.
\newblock Sun database: Large-scale scene recognition from abbey to zoo.
\newblock In \emph{IEEE Conference on Computer Vision and Pattern Recognition}, 2010.

\bibitem[Xu et~al.(2024)Xu, Xie, Tan, Huang, Howes, Sharma, Li, Ghosh, Zettlemoyer, and Feichtenhofer]{xudemystifying}
Hu Xu, Saining Xie, Xiaoqing Tan, Po-Yao Huang, Russell Howes, Vasu Sharma, Shang-Wen Li, Gargi Ghosh, Luke Zettlemoyer, and Christoph Feichtenhofer.
\newblock Demystifying {CLIP} data.
\newblock In \emph{The Twelfth International Conference on Learning Representations}, 2024.

\bibitem[Yang et~al.(2023)]{yang2023beyond}
Jia-Qi Yang et~al.
\newblock Beyond probability partitions: Calibrating neural networks with semantic aware grouping.
\newblock \emph{Advances in Neural Information Processing Systems}, 36:\penalty0 58448--58460, 2023.

\bibitem[Yoon et~al.(2024)Yoon, Yoon, Tee, Hasegawa-Johnson, Li, and Yoo]{yoon2024ctpt}
Hee~Suk Yoon, Eunseop Yoon, Joshua Tian~Jin Tee, Mark~A. Hasegawa-Johnson, Yingzhen Li, and Chang~D. Yoo.
\newblock C-{TPT}: Calibrated test-time prompt tuning for vision-language models via text feature dispersion.
\newblock In \emph{The Twelfth International Conference on Learning Representations}, 2024.

\bibitem[Yun et~al.(2019)Yun, Han, Oh, Chun, Choe, and Yoo]{yun2019cutmix}
Sangdoo Yun, Dongyoon Han, Seong~Joon Oh, Sanghyuk Chun, Junsuk Choe, and Youngjoon Yoo.
\newblock Cutmix: Regularization strategy to train strong classifiers with localizable features.
\newblock In \emph{Proceedings of the IEEE/CVF International Conference on Computer Vision (ICCV)}, pages 6023--6032, 2019.

\bibitem[Zhang et~al.(2018)Zhang, Cisse, Dauphin, and Lopez-Paz]{zhang2018mixup}
Hongyi Zhang, Moustapha Cisse, Yann~N. Dauphin, and David Lopez-Paz.
\newblock mixup: Beyond empirical risk minimization.
\newblock In \emph{International Conference on Learning Representations (ICLR)}, 2018.

\bibitem[Zhang et~al.(2020)]{zhang2020mix}
Jize Zhang et~al.
\newblock Mix-n-match: Ensemble and compositional methods for uncertainty calibration in deep learning.
\newblock In \emph{International conference on machine learning (ICML)}, 2020.

\bibitem[Zhou et~al.(2022{\natexlab{a}})Zhou, Yang, Loy, and Liu]{zhou2022cocoop}
Kaiyang Zhou, Jingkang Yang, Chen~Change Loy, and Ziwei Liu.
\newblock Conditional prompt learning for vision-language models.
\newblock In \emph{IEEE/CVF Conference on Computer Vision and Pattern Recognition (CVPR)}, 2022{\natexlab{a}}.

\bibitem[Zhou et~al.(2022{\natexlab{b}})Zhou, Yang, Loy, and Liu]{zhou2022coop}
Kaiyang Zhou, Jingkang Yang, Chen~Change Loy, and Ziwei Liu.
\newblock Learning to prompt for vision-language models.
\newblock \emph{International Journal of Computer Vision (IJCV)}, 2022{\natexlab{b}}.

\end{thebibliography}
}

\appendix
\clearpage
\setcounter{page}{1}
{
   \newpage
       \onecolumn
        \centering
        \Large
        \textbf{\thetitle}\\
        \vspace{0.5em}Supplementary Material \\
        \vspace{1.0em}
   }

\section{Proof of Proposition 1}
\label{sec:proof_prop}
We first note that for any $\vv$ and $i^\star=\arg\max_i z_i$, the confidence can be written as
\begin{align}
p_{\max}(\vv) & = \frac{\exp(z_{i^\star})}{\sum_{k=1}^K \exp(z_k)} \nonumber \\ \nonumber
& = \frac{1}{1 + \sum_{j \neq i^\star} \exp\!\big(z_j - z_{i^\star}\big)} \\
& = \frac{1}{1 + \sum_{j \neq i^\star} \exp\!\big(-\Delta z_{i^\star j}(\vv)\big)}. \nonumber
\end{align}
To obtain a universal bound, consider $\vv$ aligned with $\vt_{i^\star}$ so that $\vv^\top \vt_{i^\star}=1$ (i.e., worst). Then for any $j\neq i^\star$,
\begin{equation}
\vv^\top \vt_j \;=\; \vt_{i^\star}^\top \vt_{j} \;\le\; \max_{j\neq i^\star} \vt_{i^\star}^\top \vt_j \;=\; \mu, \nonumber
\end{equation}
hence each logit gap satisfies
\begin{equation}
\Delta \vz_{i^\star j}(\vv) 
= \alpha \big( \vv^\top \vt_{i^\star} - \vv^\top \vt_j \big)
\;\ge\; \alpha(1-\mu), 
\quad \forall j \neq i^\star. \nonumber
\end{equation}
By monotonicity of the exponential,
\begin{equation}
\exp\!\big(-\Delta z_{i^\star j}(\vv)\big) 
\;\le\; \exp\!\big(-\alpha(1-\mu)\big),
\quad \forall j \neq i^\star, \nonumber
\end{equation}
and summing over all $K-1$ competitors,
\begin{equation}
\sum_{j \neq i^\star} \exp\!\big(-\Delta z_{i^\star j}(\vv)\big)
\;\le\; (K-1)\,\exp\!\big(-\alpha(1-\mu)\big). \nonumber
\end{equation}
Substituting into the confidence expression
\begin{align}
p_{\max}(\vv) 
& = \frac{1}{1 + \sum_{j \neq i^\star} \exp\!\big(-\Delta z_{i^\star j}(\vv)\big)}\nonumber \\ 
& \;\ge\; \frac{1}{1 + (K-1)\exp\!\big(-\alpha(1-\mu)\big)}. \nonumber
\end{align}

\hfill$\Box$

\section{Exact Similarity Shifts}
\label{sec:grad_allpairs}
We derive in this section the exact similarity shifts without considering the dominant-pair approximation. Let us consider the exact gradients for both O-TPT \cite{sharifdeen2025tpt} and the proposed Huber-based regularizer:

\begin{equation}
    \nabla_{\vt_i}^{\text{O-TPT}} = 2 \sum_{k \ne i} s_{ik} \vt_k,
    \label{eq:oto_grad}
\end{equation}
and
\begin{equation}
\label{eq:hub_grad}
\nabla_{\vt_i}^{\text{Huber}} = \sum_{k \ne i} g_\delta(s_{ik}) \vt_k,
\;\text{where}\quad
g_\delta(s) =
\begin{cases}
s, & s \le \delta, \\
\delta, & s > \delta.
\end{cases}
\end{equation}

\paragraph{O-TPT \cite{sharifdeen2025tpt}.} Substituting (\ref{eq:oto_grad}) into (\ref{eq:approx}):
\begin{align}
 \label{eq:exact_otpt}
\Delta s_{ij}^{\text{O-TPT}} &\approx -\eta \left(\vt_j^\top\,\cdot 2\sum_{k \neq i}   s_{ik}\vt_k  + \vt_i^\top \cdot \, 2 \sum_{k \neq i}s_{jk}\vt_k\right)\nonumber\\ 
&\approx -2 \eta \left(\sum_{k \neq i} s_{ik}\vt_j^\top \vt_k  + \sum_{k \neq i}s_{jk}\vt_i^\top \vt_k\right)\nonumber\\
&\approx -2 \eta \left(\sum_{k \neq i} s_{ik}s_{jk}  + \sum_{k \neq j} s_{jk}s_{ik}\right).
\end{align}
By the definition of matrix multiplication,
\begin{equation}
    (SS)_{ij} \;=\; \sum_{k=1}^n S_{ik}\, S_{kj}
\;=\; \sum_{k=1}^n s_{ik}\, s_{kj}
\;=\; \sum_{k=1}^n s_{ik}\, s_{jk},\nonumber
\end{equation}
and
\begin{equation}
(SS)_{ji} \;=\; \sum_{k=1}^n s_{jk}\, s_{ik}. \nonumber
\end{equation}
If we enforce zero diagonal in the sums, (to match $k\ne i$ or $k\ne j$), as in our case, these identities still hold with the understanding that the diagonal entries do not contribute. Therefore, 
\begin{align}
\sum_{k\ne i} s_{ik}\, s_{jk} \;\equiv\; (SS)_{ij},\nonumber
\\
\sum_{k\ne j} s_{jk}\, s_{ik} \;\equiv\; (SS)_{ji}.
\label{eq:sum-to-matrix}
\end{align}
Plugging (\ref{eq:sum-to-matrix}) into (\ref{eq:exact_otpt}) yields
\begin{equation}
    \Delta s_{ij}^{\text{O-TPT}} \approx -2\eta \left((SS)_{ij}+(SS)_{ji}\right) \nonumber
\end{equation}
Considering that $S$ is symmetric, them $(SS)_{ij}=(SS)_{ji}$, and hence:
\begin{equation}
\label{eq:exact_final_otpt}
\end{equation}

\paragraph{\ours{} (Our proposed Huber-based regularizer).}
Analogously, we can plug the gradient in Eq.~(\ref{eq:hub_grad}) into Eq.~(\ref{eq:approx}):
\begin{align}
\label{eq:exact_huber}
\Delta s_{ij}^{\text{Huber}} &\approx -\eta\left(
\sum_{k\ne i} g_\delta(s_{ik})\, t_j^\top t_k
+ \sum_{k\ne j} g_\delta(s_{jk})\, t_i^\top t_k
\right)\nonumber\\ 
&\approx -\eta\left(
\sum_{k\ne i} g_\delta(s_{ik})\, s_{jk}
+ \sum_{k\ne j} g_\delta(s_{jk})\, s_{ik}
\right).
\end{align}
Let $G = g_\delta(S)$ be the element-wise application of the Huber gradient function to $S$,
with $G_{ij} = g_\delta(s_{ij})$ and $G_{ii} = 0$.
Then the expression
\begin{equation}
\sum_{k\ne i} g_\delta(s_{ik})\, s_{jk}
\quad\text{and}\quad
\sum_{k\ne j} g_\delta(s_{jk})\, s_{ik} \nonumber
\end{equation}
can be written as matrix products:
\begin{equation}
(GS)_{ij} = \sum_k G_{ik} S_{kj} = \sum_k g_\delta(s_{ik})\, s_{jk},\nonumber
\end{equation}
and similarly for $(GS)_{ji}$.
Thus, Eq. \ref{eq:exact_huber} becomes
\begin{equation}
\Delta s_{ij}^{\text{Huber}} \approx -\eta\left[(GS)_{ij} + (GS)_{ji}\right].\nonumber
\label{eq:huber-matrix}
\end{equation}
If $S$ is symmetric, then $(GS)_{ij} = (GS)_{ji}$, yielding:
\begin{equation}
\Delta s_{ij}^{\text{Huber}} \approx -2\eta\, (GS)_{ij}.
\label{eq:huber-exact_final}
\end{equation}

\paragraph{Implications for confidence bounds.} The exact similarity shift derived above provides the structural foundation for Corollary~\ref{cor:confidence_increase}. Specifically, the matrix form (Eqs. (\ref{eq:exact_final_otpt}) and (\ref{eq:huber-exact_final}))
\begin{equation}
\Delta s_{ij}^{\text{O-TPT}} \approx -4\eta (S^2)_{ij}
\quad\text{vs.}\quad
\Delta s_{ij}^{\text{Huber}} \approx -2\eta (GS)_{ij}\nonumber
\end{equation}
reveals that O-TPT applies stronger gradients to high-similarity regions, leading to a more pronounced reduction in worst-case similarity $\mu = \max_{i \ne j} s_{ij}$. Since the softmax confidence lower bound $p_{\max}(\vv)$ depends inversely on $\mu$ (see Proposition~\ref{prop:bound}), the sharper contraction under O-TPT yields a strictly higher confidence bound. In other words, the stronger second-order structure of O-TPT translates directly into more confident predictions, especially in regimes where prototype overlap is high.

\section{Datasets}
\label{sec:data}
As stated in the main paper, to ensure a comprehensive and comparable evaluation, we follow established practices in recent prompt tuning literature \cite{shu2022tpt, yoon2024ctpt, sharifdeen2025tpt, zhou2022coop}, assessing our approach on 11 widely studied datasets spanning diverse visual domains. These datasets have been consistently adopted in prior works to probe generalization, domain sensitivity, and semantic diversity, making them a strong basis for evaluating prompt-based methods. Detailed dataset statistics and descriptions are provided in~\Cref{tab:datasets}.
\begin{table*}[h]
    \caption{\textbf{Description of the datasets.} Number of classes, number of images in the test set, and brief description of the type of images.}
    \label{tab:datasets}
    \centering
    \begin{tabular}{l c c l}
    \toprule
    & Num. classes & Num. images & Description \\
    \midrule
        ImageNet~\cite{5206848} & 1,000 & 50,000 & Large-scale dataset of natural images.\\
        DTD~\cite{cimpoi14describing} & 47 & 1,692 & Diverse visual textures. \\
        Flowers102~\cite{Nilsback08} & 102 & 2,463 & Various species of flowers.\\
        Food101~\cite{bossard14} & 101 & 30,300 & Popular foods from around the world.\\
        SUN397~\cite{xiao2010sun} & 397 & 19,850 & Scene recognition dataset.\\
        FGVC Aircraft~\cite{maji13fine-grained} & 100 & 3,333 & Images of aircrafts, classified by model.\\
        OxfordPets~\cite{parkhi12a} & 37 & 3,669 & Images of pets, classified by breed. \\
        Caltech101~\cite{li2006one} & 101 & 2,465 & Natural images from everyday objects.\\
        UCF101~\cite{Soomro2012UCF101AD} & 101 & 3,783 & Human action recognition.\\
        EuroSAT~\cite{helber2017eurosat} & 10 & 8,100 & Satellite images of different types of land use.\\
        StanfordCars~\cite{KrauseStarkDengFei-Fei_3DRR2013} & 196 & 8,041 & Cars classified by brand and model.\\
    \midrule
        ImageNet-A~\cite{imagenet_a} & 200 & 7,500 & Natural images, challenging for most classifiers. \\
        ImageNet-v2~\cite{pmlr-v97-recht19a} & 1,000 & 10,000 & Resampled ImageNet test set to verify robustness. \\
        ImageNet-R~\cite{hendrycks2021many} & 200 & 30,000 & Artistic renditions of ImageNet classes. \\
        ImageNet-Sketch~\cite{wang2019learning} & 1,000 & 50,889 & Drawn images from the ImageNet classes. \\
    \bottomrule
    \end{tabular}
\end{table*}
\section{Additional Implementation Details}
\label{sec:implementation}

We normalize the cosine similarities $\tilde{\mS} = \frac{\mS-\mS_\text{min}}{\mS_\text{max} - \mS_\text{min}}$, to account for dataset-dependent class correlations. We choose the margin $\delta$ to be the $20$-th percentile of $\tilde \mS$. We chose the value of the weight of the regularizer $\lambda=30$ for all experiments, except for the distribution shift experiments, where, similarly to previous work~\cite{sharifdeen2025tpt}, we choose a smaller value of $\lambda=14$.
In order to keep an optimal scaling factor across all datasets, we use $\lambda \frac{\vert \mathcal L_\text{SoC}\vert}{\vert \mathcal L_\text{TPT}\vert}$, ensuring a constant optimal value.
For C-TPT~\cite{yoon2024ctpt} and O-TPT~\cite{sharifdeen2025tpt}, we use the original values of $\lambda$ reported in the original works.
All experiments were performed on NVIDIA A100 GPUs with 80 GB of memory.
\section{Additional Experiments}
\label{sec:additional_experiments}
\mypar{Stability to initialization.} In Tab.~\ref{tab:std_analysis}, we apply C-TPT, O-TPT, and our method across six datasets on three seeds (we chose six datasets with relatively small size to reduce the compute load), and report the standard deviation of the reported accuracy and calibration error. The lower standard deviation of both metrics indicated that our method is more stable to random initializations.
\begin{table}[h!]
    \caption{\textbf{Standard deviations across three seeds:} accuracy and ECE across 6 datasets compared with state-of-the-art.}
    \label{tab:std_analysis}
    \centering
    \begin{tabular}{l l c c c c c c c}
        \toprule
         & & DTD
         & Flowers
         & Aircraft
         & Pets
         & Caltech
         & UCF101
         & Average \\
        \midrule
        \multirow{3}{*}{\rotatebox{90}{\textbf{Acc.}}} & C-TPT &
            0.12 &
            0.29 &
            0.38 &
            0.09 &
            0.18 &
            0.08 &
            0.19 \\
        & O-TPT &
            0.06 &
            0.15 &
            0.23 &
            0.07 &
            0.06 &
            0.14 &
            0.12 \\
        & \our \textbf{\ours{} (\textit{Ours})} &
          \our  0.16 &
          \our  0.06 &
          \our  0.03 &
          \our  0.04 &
          \our  0.13 &
          \our  0.03 &
          \our  0.08 \\
        \midrule
        \multirow{3}{*}{\rotatebox{90}{\textbf{ECE}}} & C-TPT &
            0.09 &
            0.30 &
            0.42 &
            0.04 &
            0.23 &
            0.04 &
            0.19 \\
        & O-TPT &
            0.15 &
            0.20 &
            0.20 &
            0.13 &
            0.22 &
            0.08 &
            0.16 \\
        & \our \textbf{\ours{} (\textit{Ours})} &
           \our 0.15 &
           \our 0.04 &
           \our 0.07 &
           \our 0.14 &
          \our  0.17 &
           \our 0.22 &
           \our 0.13 \\
        \bottomrule
    \end{tabular}
    
\end{table}

\mypar{Normalization strategies} Tab.~\ref{tab:ablation_norm} shows an ablation with three different normalizations for \ours{}, justifying our choice for the $\tilde{\mS} = \frac{\mS-\mS_\text{min}}{\mS_\text{max} - \mS_\text{min}}$ normalization.
\begin{table}[t]
    \centering
    \caption{\textbf{Accuracy and calibration error of \ours{} using three different normalizations.} Average performed over eight datasets. $\tilde \mS_1 = \frac{\mS-\mS_\text{min}}{\mS_\text{max} - \mS_\text{min}}$, $\tilde \mS_2 = \frac{\mS}{\mS_\text{max}}$, and $\tilde \mS_3 =\mS-\mS_\text{min}$.}
    \label{tab:ablation_norm}
        \centering
        \begin{tabular}{l c c c}
        \toprule
            & $\tilde \mS_1$ & $\tilde \mS_2$ & $\tilde \mS_3$ \\
        \midrule
            Accuracy & 70.3 & 70.7 & 69.2 \\
            ECE & 5.3 & 7.3 & 5.5 \\
        \bottomrule
        \end{tabular}
\end{table}

\mypar{Computational efficiency.}
The computational complexity of our \ours{} implementation is $O(n^2)$, while O-TPT has $O(n^3)$ complexity, due to the Householder transform, where $n$ is the number of classes. Tab.~\ref{tab:complexity} shows the FLOPS of \ours{} and O-TPT for varying number of classes, which aligns with the lower computational complexity.
\begin{table}[h!]
    \centering
    \caption{\textbf{Computational efficiency for varying number of classes}. MFLOPS of O-TPT and \ours{} for number of classes ranging from the smallest number (EuroSAT) to the highest number (ImageNet). Values correspond to the computational efficiency of computing the $\mathcal L _\text{O-TPT}$ and $\mathcal L _\text{\ours{}}$ losses (i.e. not including the $\mathcal L _\text{TPT}$ term).}
    \label{tab:complexity}
    \begin{tabular}{l c c c}
    \toprule
    & 10 & 100 & 1000 \\
    \midrule
    O-TPT & 0.2 & 20.4 & 6537.0 \\
    \our \textbf{\ours{} (\textit{Ours})} &
        \our 0.2 &
        \our 15.4 &
        \our 1536.0 \\
    \bottomrule
    \end{tabular}
\end{table}

\mypar{Detailed results.} Tab.~\ref{tab:vitl_2step} shows the per-dataset results corresponding to the two-step updates shown in Fig.~\ref{fig:2steps}. The gaps in calibration error between \ours{} and O-TPT become very apparent, with certain datasets showing differences in ECE of 10.1 (DTD) and 26.1 (EuroSAT).
Tab.~\ref{tab:vitl_sals} shows the per-dataset results of applying SaLS~\cite{murugesan2024robust} on top of C-TPT, O-TPT, and \ours{} (Fig.~\ref{fig:sals}). Because SaLS only scales the logits, the accuracy of the methods does not change, but the calibration error drops across all methods. As expected, as the models get better calibrated, the gap between methods is reduced.
\begin{table*}[t]
    \caption{\textbf{Applying two gradient updates.} Accuracy and ECE across 11 datasets compared with two baseline methods, when applying two gradient updates. Green and red indicate positive and negative changes with respect to O-TPT respectively.}
    \label{tab:vitl_2step}
    \centering
    \resizebox{\linewidth}{!}{
    \begin{tabular}{l l c c c c c c c c c c c c}
        \toprule &
         & ImgNet
         & DTD
         & Flowers
         & Food101
         & SUN397
         & Aircraft
         & Pets
         & Caltech
         & UCF101
         & EuroSAT
         & Cars
         & Average \\
        \midrule
        \multirow{3}{*}{\rotatebox{90}{\textbf{Acc.}}} & C-TPT &
            75.2 &
            54.4 &
            76.5 &
            89.0 &
            70.7 &
            30.0 &
            94.1 &
            95.4 &
            74.8 &
            50.7 &
            76.5 &
            71.6 \\
        & O-TPT &
            72.8 &
            54.1 &
            76.0 &
            88.7 &
            69.0 &
            27.5 &
            93.2 &
            95.4 &
            73.9 &
            50.2 &
            77.1 &
            70.7 \\
        & \our \textbf{\ours{} (\textit{Ours})} &
            \our 72.0\imp{1.4} &
            \our 55.6\imp{+1.5} &
            \our 77.9\imp{+1.9} &
            \our 88.9\imp{+0.2} &
            \our 70.1\imp{+1.1} &
            \our 27.7\imp{+0.2} &
            \our 94.1\imp{+0.9} &
            \our 95.7\imp{+2.0} &
            \our 75.9\imp{+5.2} &
            \our 55.4\imp{-0.7} &
            \our 76.4\wor{+1.4} &
            \our 71.8\imp{+1.3} \\
        \midrule
        \multirow{3}{*}{\rotatebox{90}{\textbf{ECE}}} & C-TPT &
            14.4 &
            26.6 &
            13.3 &
            4.7 &
            18.8 &
            28.9 &
            1.0 &
            2.0 &
            16.1 &
            29.2 &
            2.6 &
            14.4 \\
        & O-TPT &
            7.8 &
            21.2 &
            8.6 &
            3.9 &
            10.0 &
            22.8 &
            0.9 &
            1.1 &
            11.5 &
            29.5 &
            1.4 &
            10.8 \\
        & \our \textbf{\ours{} (\textit{Ours})} &
            \our 10.3\wor{+2.5} &
            \our 11.1\imp{-10.1} &
            \our 6.6\imp{-2.0} &
            \our 2.0\imp{-1.9} &
            \our 8.0\imp{-2.0} &
            \our 20.8\imp{-2.0} &
            \our 1.1\wor{+0.2} &
            \our 1.1\imp{-0.0} &
            \our 6.8\imp{-4.7} &
            \our 3.4\imp{-26.1} &
            \our 1.4\imp{-0.0} &
            \our 6.6\imp{-4.2} \\
        \bottomrule
    \end{tabular}}
\end{table*}
\begin{table*}[t]
    \caption{\textbf{Applying SaLS \cite{murugesan2024robust} on ViT-L/14 backbone.} Accuracy and ECE across 11 datasets compared with two baseline methods, when applying SaLS on top of the TPT-based method. Green and red indicate positive and negative changes with respect to O-TPT respectively.}
    \label{tab:vitl_sals}
    \centering
    \resizebox{\linewidth}{!}{
    \begin{tabular}{l l c c c c c c c c c c c c}
        \toprule
         & & 
         ImgNet
         & DTD
         & Flowers
         & Food101
         & SUN397
         & Aircraft
         & Pets
         & Caltech
         & UCF101
         & EuroSAT
         & Cars
         & Average \\
        \midrule
        \multirow{3}{*}{\rotatebox{90}{\textbf{Acc.}}} & C-TPT &
            75.0 &
            55.1 &
            76.5 &
            88.9 &
            70.1 &
            30.9 &
            94.1 &
            95.5 &
            75.2 &
            54.0 &
            77.5 &
            72.1 \\
        & O-TPT &
            73.2 &
            54.6 &
            76.4 &
            88.6 &
            68.9 &
            30.0 &
            93.8 &
            95.3 &
            74.5 &
            53.6 &
            76.7 &
            71.4 \\
        & \our \textbf{\ours{} (\textit{Ours})} &
           \our74.5\imp{+1.3} &
           \our54.4\wor{-0.2} &
           \our77.0\imp{+0.6} &
           \our88.9\imp{+0.3} &
           \our69.5\imp{+0.6} &
           \our30.9\imp{+0.9} &
           \our93.9\imp{+0.1} &
           \our95.6\imp{+0.3} &
           \our74.9\imp{+0.4} &
           \our58.3\imp{+4.7} &
           \our77.0\imp{+0.3} &
           \our72.3\imp{+0.9} \\
        \midrule
        \multirow{3}{*}{\rotatebox{90}{\textbf{ECE}}} & C-TPT &
            8.1 &
            13.7 &
            9.1 &
            1.5 &
            8.8 &
            17.9 &
            1.6 &
            1.6 &
            8.7 &
            13.4 &
            2.2 &
            7.9 \\
        & O-TPT &
            3.8 &
            9.9 &
            5.4 &
            1.5 &
            3.6 &
            13.6 &
            2.1 &
            2.7 &
            6.2 &
            13.7 &
            3.6 &
            6.0 \\
        & \our \textbf{\ours{} (\textit{Ours})} &
           \our 6.1\wor{+2.3}&
           \our 9.8\imp{-0.1} &
           \our 4.7\imp{-0.7} &
           \our 1.6\wor{+0.1} &
           \our 5.1\wor{+1.5} &
           \our 12.2\imp{-1.4} &
           \our 1.2\imp{-0.9} &
           \our 1.8\imp{-0.9} &
           \our 5.6\imp{-0.6} &
           \our 3.4\imp{-10.3} &
           \our 2.8\imp{-0.8} &
           \our 4.9\imp{-1.1} \\
        \bottomrule
    \end{tabular}}
\end{table*}

\begin{figure*}
    \centering
    \begin{subfigure}[b]{0.24\linewidth}
        \centering
        \includegraphics[width=\linewidth]{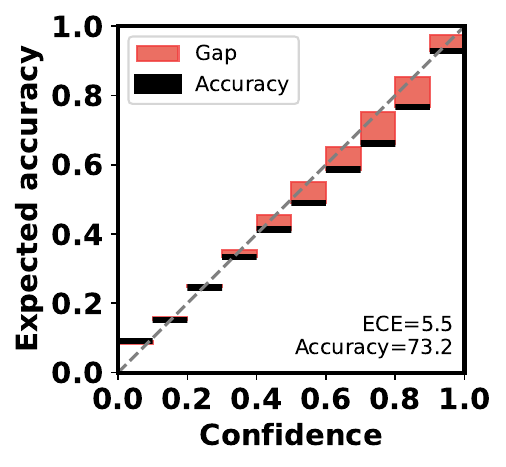}
        \caption{ImageNet O-TPT}
    \end{subfigure}
    \begin{subfigure}[b]{0.24\linewidth}
        \centering
        \includegraphics[width=\linewidth]{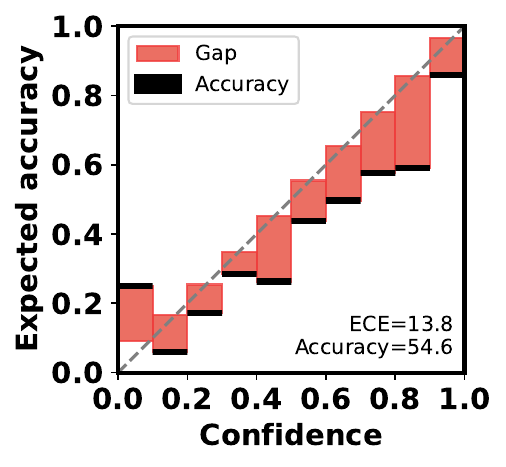}
        \caption{DTD O-TPT}
    \end{subfigure}
    \begin{subfigure}[b]{0.24\linewidth}
        \centering
        \includegraphics[width=\linewidth]{figures/calibration/calibration_O-TPT_Flowers.pdf}
        \caption{Flowers102 O-TPT}
    \end{subfigure}
    \begin{subfigure}[b]{0.24\linewidth}
        \centering
        \includegraphics[width=\linewidth]{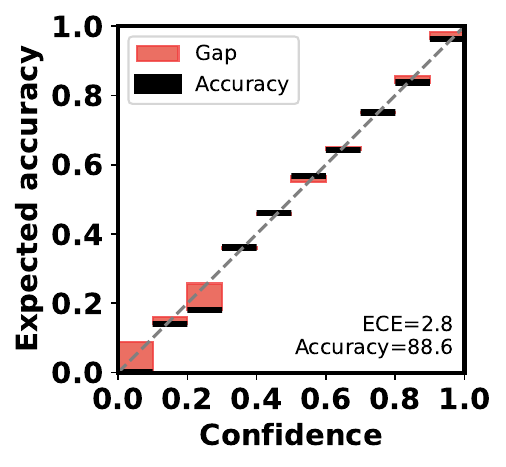}
        \caption{Food101 O-TPT}
    \end{subfigure} \\
    \begin{subfigure}[b]{0.24\linewidth}
        \centering
        \includegraphics[width=\linewidth]{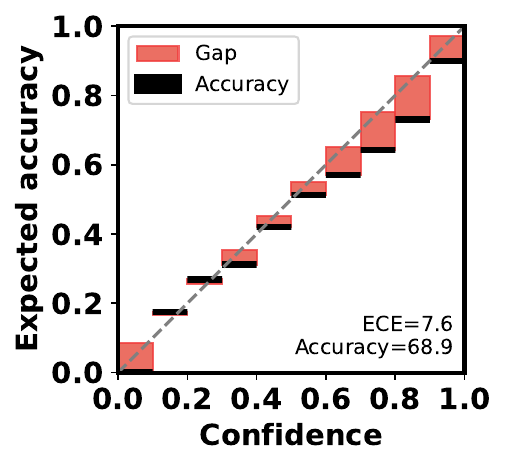}
        \caption{SUN397 O-TPT}
    \end{subfigure}
    \begin{subfigure}[b]{0.24\linewidth}
        \centering
        \includegraphics[width=\linewidth]{figures/calibration/calibration_O-TPT_Aircraft.pdf}
        \caption{Aircraft O-TPT}
    \end{subfigure}
    \begin{subfigure}[b]{0.24\linewidth}
        \centering
        \includegraphics[width=\linewidth]{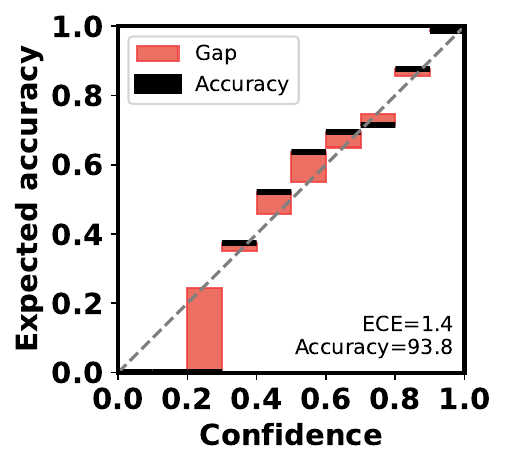}
        \caption{OxfordPets O-TPT}
    \end{subfigure}
    \begin{subfigure}[b]{0.24\linewidth}
        \centering
        \includegraphics[width=\linewidth]{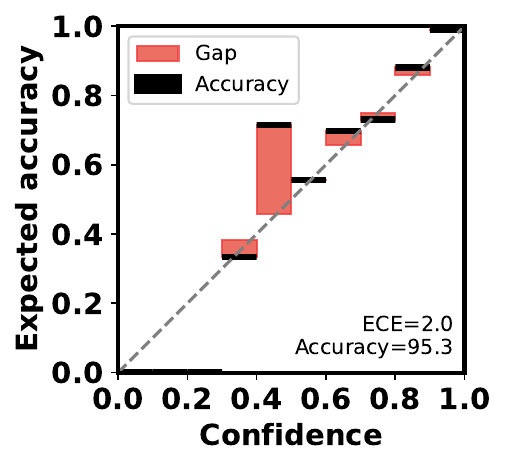}
        \caption{Caltech101 O-TPT}
    \end{subfigure} \\
    \begin{subfigure}[b]{0.24\linewidth}
        \centering
        \includegraphics[width=\linewidth]{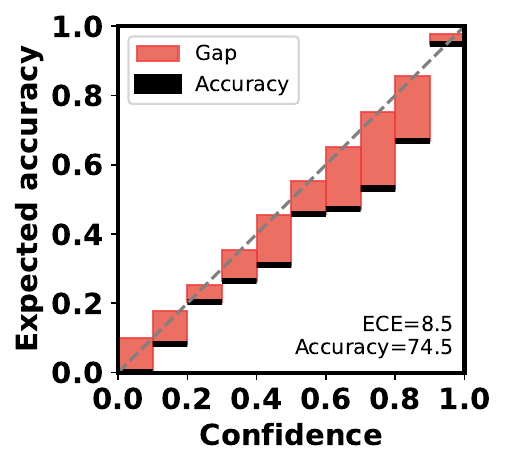}
        \caption{UCF101 O-TPT}
    \end{subfigure}
    \begin{subfigure}[b]{0.24\linewidth}
        \centering
        \includegraphics[width=\linewidth]{figures/calibration/calibration_O-TPT_EuroSAT.pdf}
        \caption{EuroSAT O-TPT}
    \end{subfigure}
    \begin{subfigure}[b]{0.24\linewidth}
        \centering
        \includegraphics[width=\linewidth]{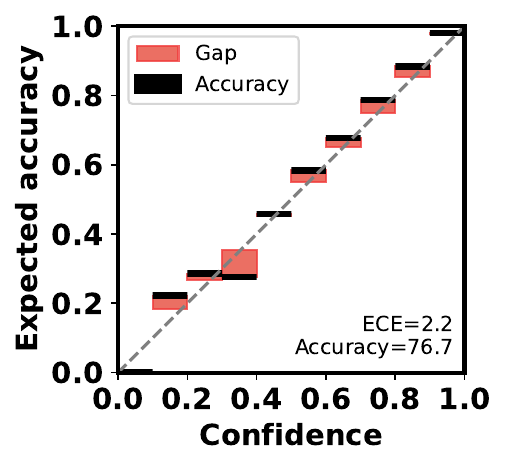}
        \caption{Cars O-TPT}
    \end{subfigure}
    \caption{\textbf{Reliability plots for O-TPT.} Calibration error across all 11 datasets for O-TPT.}
    \label{fig:reliability_otpt}
\end{figure*}

\begin{figure*}
    \centering
    \begin{subfigure}[b]{0.24\linewidth}
        \centering
        \includegraphics[width=\linewidth]{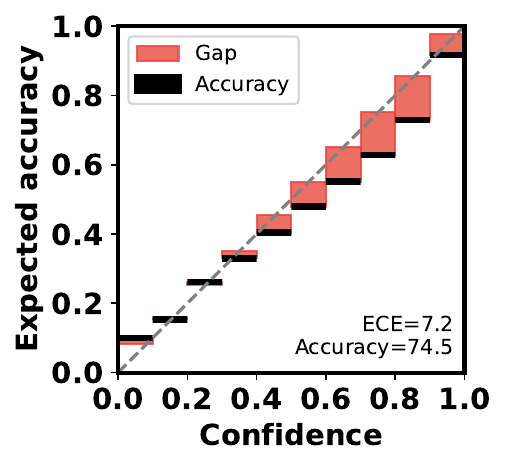}
        \caption{ImageNet \ours{}}
    \end{subfigure}
    \begin{subfigure}[b]{0.24\linewidth}
        \centering
        \includegraphics[width=\linewidth]{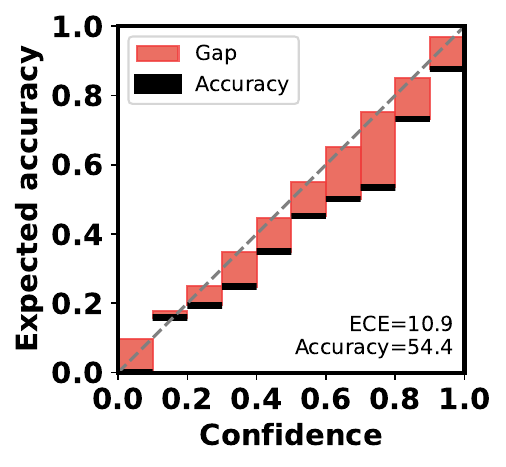}
        \caption{DTD \ours{}}
    \end{subfigure}
    \begin{subfigure}[b]{0.24\linewidth}
        \centering
        \includegraphics[width=\linewidth]{figures/calibration/calibration_Ours_Flowers.pdf}
        \caption{Flowers102 \ours{}}
    \end{subfigure}
    \begin{subfigure}[b]{0.24\linewidth}
        \centering
        \includegraphics[width=\linewidth]{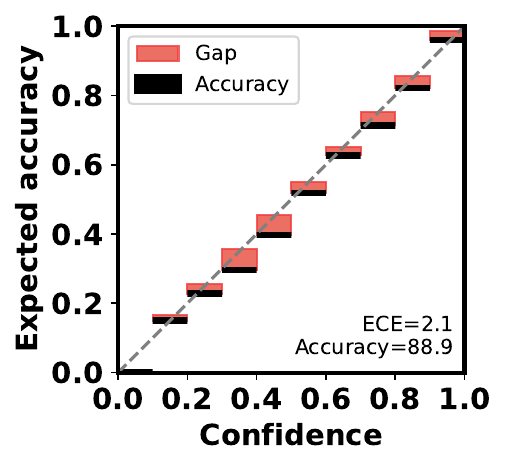}
        \caption{Food101 \ours{}}
    \end{subfigure} \\
    \begin{subfigure}[b]{0.24\linewidth}
        \centering
        \includegraphics[width=\linewidth]{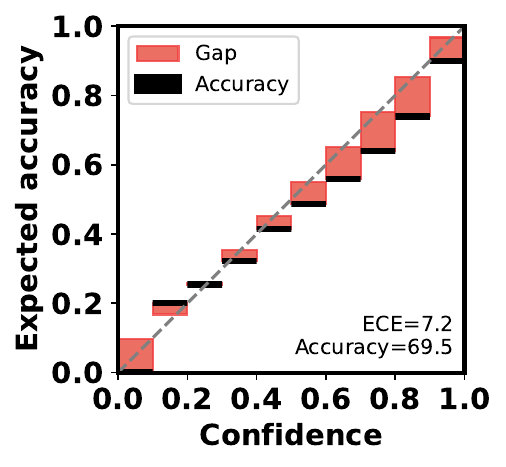}
        \caption{SUN397 \ours{}}
    \end{subfigure}
    \begin{subfigure}[b]{0.24\linewidth}
        \centering
        \includegraphics[width=\linewidth]{figures/calibration/calibration_Ours_Aircraft.pdf}
        \caption{Aircraft \ours{}}
    \end{subfigure}
    \begin{subfigure}[b]{0.24\linewidth}
        \centering
        \includegraphics[width=\linewidth]{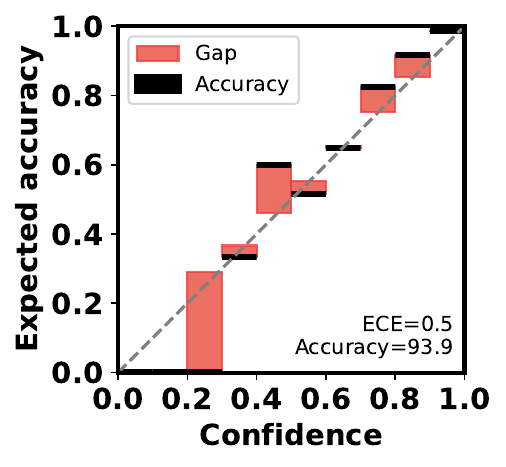}
        \caption{OxfordPets \ours{}}
    \end{subfigure}
    \begin{subfigure}[b]{0.24\linewidth}
        \centering
        \includegraphics[width=\linewidth]{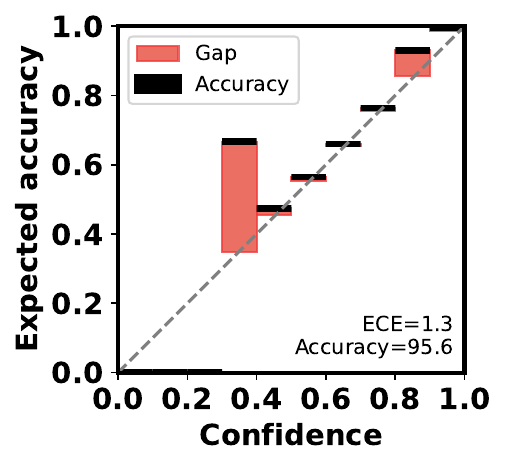}
        \caption{Caltech101 \ours{}}
    \end{subfigure} \\
    \begin{subfigure}[b]{0.24\linewidth}
        \centering
        \includegraphics[width=\linewidth]{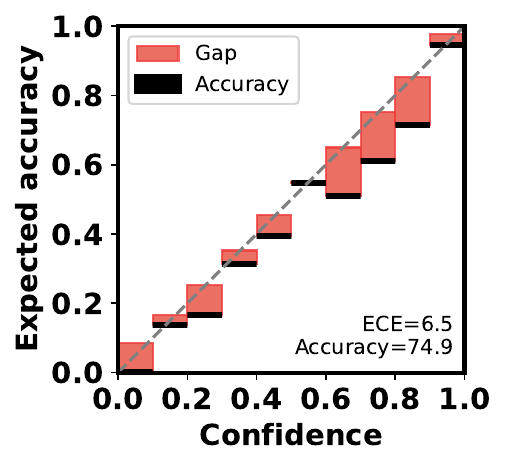}
        \caption{UCF101 \ours{}}
    \end{subfigure}
    \begin{subfigure}[b]{0.24\linewidth}
        \centering
        \includegraphics[width=\linewidth]{figures/calibration/calibration_Ours_EuroSAT.pdf}
        \caption{EuroSAT \ours{}}
    \end{subfigure}
    \begin{subfigure}[b]{0.24\linewidth}
        \centering
        \includegraphics[width=\linewidth]{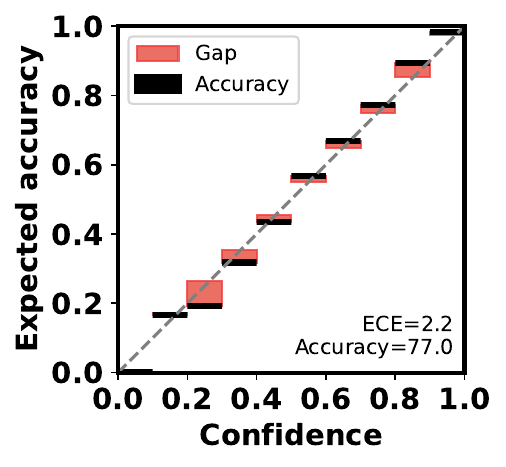}
        \caption{Cars \ours{}}
    \end{subfigure}
    \caption{\textbf{Reliability plots for \ours{}.} Calibration error across all 11 datasets for \ours{}.}
    \label{fig:reliability_ours}
\end{figure*}

\mypar{Reliability plots.} Fig.~\ref{fig:reliability_otpt} and Fig.~\ref{fig:reliability_ours} shows the reliability plots across all datasets for O-TPT~\cite{sharifdeen2025tpt} and \ours{}, respectively. While EuroSAT is the most notable example, where calibration errors mostly vanish from O-TPT to \ours{}, the difference in calibration can also clearly be seen from the reliability plots in other datasets such as DTD, Flowers102, Aircraft, and UCF101. Analyzing the plots across all datasets confirms the overall tendency of O-TPT to be overconfident on its predictions.

\mypar{Adaptive calibration error.} The adaptive calibration error (ACE) is an alternative metric to the ECE, which defines the bins to not be equally spaced (as in ECE), but to all contain the same number of samples. Formally, the number of samples in each bin $n_i=\frac{M}{\vert B \vert}, \forall i\in\{1,\ldots,\vert B \vert\}$, where $M$ is the size of the test set, and $B$ represents the bins. Tab.~\ref{tab:vitl_ace} shows the results of the ACE corresponding to the ViT-L/14 (i.e. an extension of Tab.~\ref{tab:vitl}).
\begin{table*}[t]
    \caption{\textbf{Adaptive calibration error on ViT-L/14.} ACE across 11 datasets compared with other baseline methods. Green and red indicate positive and negative changes with respect to O-TPT respectively.}
    \label{tab:vitl_ace}
    \centering
    \resizebox{\linewidth}{!}{
    \begin{tabular}{l l c c c c c c c c c c c c}
        \toprule &
         & ImgNet
         & DTD
         & Flowers
         & Food101
         & SUN397
         & Aircraft
         & Pets
         & Caltech
         & UCF101
         & EuroSAT
         & Cars
         & Average \\
        \midrule
        \multirow{5}{*}{\rotatebox{90}{\textbf{ACE}}} & Zero-shot &
            2.9 &
            10.2 &
            4.5 &
            1.7 &
            3.3 &
            10.3 &
            2.4 &
            3.7 &
            5.7 &
            7.2 &
            3.9 &
            5.1 \\
        & TPT &
            14.8 &
            24.8 &
            15.0 &
            6.2 &
            17.1 &
            28.8 &
            3.5 &
            2.3 &
            17.6 &
            25.8 &
            7.1 &
            14.8 \\
        & C-TPT &
            10.5 &
            18.3 &
            11.1 &
            3.3 &
            13.4 &
            21.4 &
            1.0 &
            0.9 &
            11.1 &
            17.3 &
            1.4 &
            10.0 \\
        & O-TPT &
            5.5 &
            13.7 &
            7.2 &
            2.8 &
            7.5 &
            16.8 &
            1.2 &
            1.2 &
            8.5 &
            17.8 &
            2.2 &
            7.7 \\
        & \our \textbf{\ours{} (\textit{Ours})} &
            \our7.2\wor{+1.7} &
            \our10.9\imp{+2.8} &
            \our5.5\imp{+1.7} &
            \our1.9\imp{+0.9} &
            \our7.1\imp{+0.4} &
            \our12.7\imp{+4.1} &
            \our0.5\imp{+0.7} &
            \our1.5\wor{-0.3} &
            \our6.3\imp{+2.2} &
            \our3.5\imp{+14.3} &
            \our2.2\imp{+0.0} &
            \our5.4\imp{-2.3} \\
        \bottomrule
    \end{tabular}
    }
\end{table*}

\begin{figure}
    \centering
    \begin{subfigure}[b]{0.22\linewidth}
        \centering
        \includegraphics[width=\linewidth]{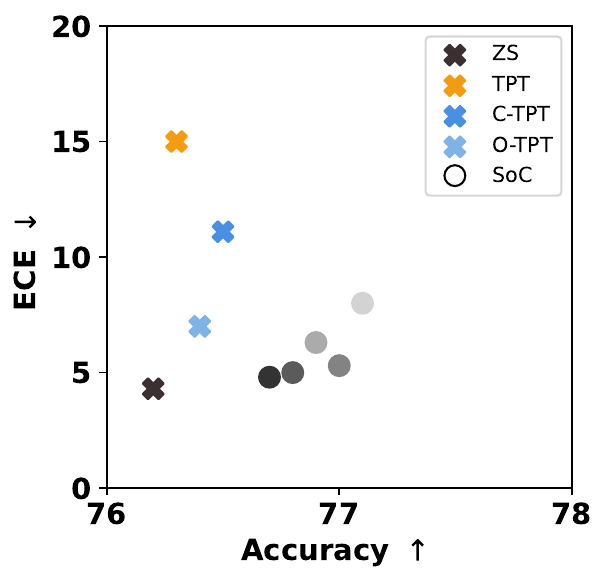}
        \caption{Flowers}
    \end{subfigure}
    \begin{subfigure}[b]{0.22\linewidth}
        \centering
        \includegraphics[width=\linewidth]{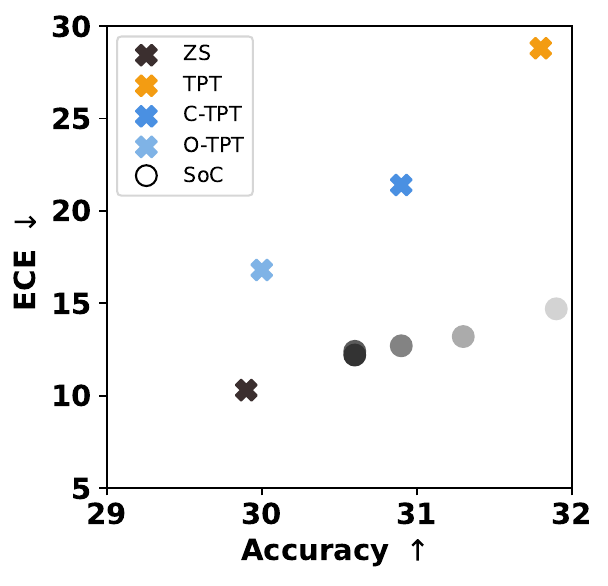}
        \caption{Aircraft}
    \end{subfigure}
    \begin{subfigure}[b]{0.22\linewidth}
        \centering
        \includegraphics[width=\linewidth]{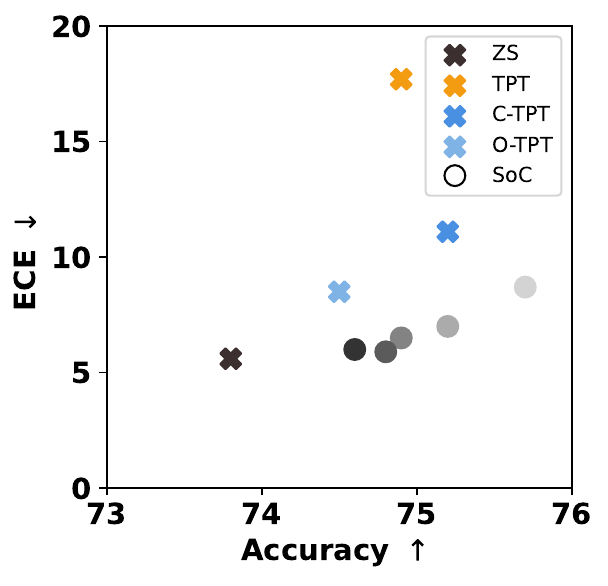}
        \caption{UCF101}
    \end{subfigure}
    \begin{subfigure}[b]{0.26\linewidth}
        \centering
        \includegraphics[width=\linewidth]{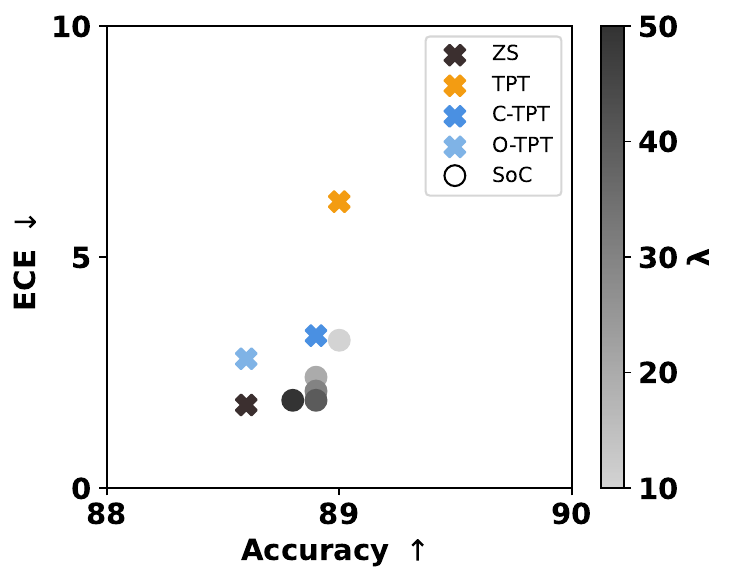}
        \caption{Food101}
    \end{subfigure}
    \caption{\textbf{Ablation on the value of $\lambda$.} Accuracy and calibration error for multiple values of the regularization term $\lambda$ of \ours{} for the Flowers, Aircraft, UCF101, and Food101 datasets.}
    \label{fig:lambda_ablation}
\end{figure}

\mypar{Ablation on $\lambda$.} The weight $\lambda$ impacts the importance given to the regularization term relative to the TPT loss term, and acts as a tradeoff between the accuracy and the calibration. Higher values of $\lambda$ will give more importance to the calibration, whereas lower values will favor better accuracy. Fig.~\ref{fig:lambda_ablation} shows how the model performs in the accuracy-calibration space for different values of $\lambda$ ranging from $10$ to $50$. Most points live in the lower-right side of O-TPT indicating better calibration and accuracy regardless of the chosen value of $\lambda$.

\end{document}